\documentclass[runningheads]{llncs}

\usepackage{eccv}

\usepackage{eccvabbrv}

\usepackage{graphicx}
\usepackage{booktabs}
\usepackage[accsupp]{axessibility}  
\usepackage{algorithm}
\usepackage{algpseudocode}
\usepackage{censor}

\usepackage{amsmath,amssymb}
\usepackage{orcidlink}
\usepackage{xcolor}
\definecolor{mygray}{gray}{.9}
\definecolor{Mygray}{gray}{0.35}
\usepackage{color, colortbl}
\begin{document}

\title{CARA: Collision-Aware Resolution Adaptation for Multiresolution Hash Encoding Based Image Fitting} 

\titlerunning{CARA}

\author{Linfeng Ye\inst{1, 3}\orcidlink{0009-0009-2355-1773} \and
Zhixiang Chi\inst{1}\orcidlink{0000-0003-4560-4986} \and
Shayan Mohajer Hamidi\inst{2}\orcidlink{0000-0001-8321-7130} \and
En-hui Yang \inst{3}\orcidlink{0000-0003-1504-7584} \and 
Konstantinos N.~Plataniotis\inst{1}\orcidlink{0000-0003-3647-5473}}

\authorrunning{L.~Ye et al.}

\institute{
\mbox{$^{1}$University of Toronto \quad $^{2}$Stanford University \quad $^{3}$University of Waterloo}\\
\mbox{\email{linfeng.ye@mail.utoronto.ca; \{zhxchi;kostas\}@ece.utoronto.ca;}}\\
\mbox{\email{smohajer@stanford.edu; ehyang@uwaterloo.ca}}
}

\maketitle

\begin{abstract}

Multiresolution hash encodings have recently enabled fast and high-fidelity implicit neural representations by storing multi-scale features in fixed-size hash tables along a geometric resolution schedule. However, the standard design is data-agnostic: different resolution levels receive identical hash-table capacity despite large differences in image frequency content. As a result, some levels experience severe hash collisions while others underutilize parameters, leading to inefficient capacity allocation.
To address this issue, we propose \textbf{C}ollision-\textbf{A}ware \textbf{R}esolution \textbf{A}daptation (CARA), a method that assigns per-level resolutions by balancing the effective information load across hash levels. This adaptive allocation reduces capacity bottlenecks and improves parameter efficiency.
In addition, we introduce an invertible pixel-shuffle transform that reduces hash load factors by redistributing spatial information, thereby mitigating collision-induced information loss without enlarging the hash tables.
To support evaluation on extremely high-resolution data, we also curate, to the best of our knowledge, the first uncompressed whole-slide image dataset for academic research.
Experiments on Kodak images, gigapixel natural images, and raw whole-slide images demonstrate that CARA consistently improves the fidelity–parameter trade-off. Our method matches state-of-the-art performance while using only $27.76\%$ of the parameters, and achieves up to $6.11 dB$ PSNR improvement at comparable parameter counts. Code is provided in the supplementary.
\end{abstract}
\section{Introduction}
\label{sec:intro}
Implicit neural representations (INRs) \cite{sitzmann2020implicit, tancik2020fourier} have reshaped signal processing and computer vision by replacing discrete grid-based representations with continuous coordinate-based functional mappings. By parameterizing signals using deep neural networks  \cite{chen2019learning, chen2021nerv, chen2023factor}, INRs provide a flexible and expressive framework for representing images. However, conventional multi-layer perceptrons (MLPs) often suffer from a spectral bias toward low-frequency components, which hinders the recovery of fine-grained, high-frequency details \cite{mildenhall2021nerf}. While architectural techniques such as Fourier feature mappings and periodic activation functions partially alleviate this issue \cite{tancik2020fourier, liu2024finer, sitzmann2020implicit}, focus has shifted toward hybrid grid-based representations, such as Instant-NGP \cite{muller2022instant, takikawa2023compact}. These methods accelerate optimization via multiresolution hash encoding, where spatial coordinates are mapped into a hierarchy of learnable hash tables and subsequently decoded by a lightweight MLP.

%

Despite their empirical success, existing multiresolution hash encoding-based methods suffer from two fundamental limitations as discussed in the sequel:

(\textbf{i}) \textit{Data-agnostic resolution schedules:} They largely rely on data-agnostic encoding structures \cite{muller2022instant, takikawa2023compact, tang2021octfield}, which require manual hyperparameter tuning.
In particular, they employ fixed, hand-crafted geometric resolution schedules that are shared across all images, regardless of spatial or temporal characteristics. 
However, images exhibit highly non-uniform information distributions across frequency bands (see \Cref{Sec:method-CARA}). As a result, fixed resolution schedules may allocate excessive parameters to frequency regions with limited information content, while under-allocating capacity to information-dense regions that dominate perceptual error.

(\textbf{ii}) \textit{Information degradation due to hash collisions:} Multiresolution hash encodings inherently suffer from hash collisions, where multiple grid vertices are mapped to the same table entry. The impact of such collisions varies significantly across resolution levels and depends on the underlying information density. Treating resolution assignment and collision behavior as independent design factors therefore leads to suboptimal parameter utilization.

We address both limitations with a unified framework. To address (i), we adapt per-level resolutions to balance the information load across frequency bands, preventing over-parameterization in sparse bands and under-capacity in information-dense bands, thereby significantly improving parameter efficiency. To address (ii), we apply an invertible pixel-shuffle transform that reduces hash collision by lowering the effective load factor before hashing.



We refer to the resulting framework as \textbf{c}ollision-\textbf{a}ware \textbf{r}esolution \textbf{a}daptation (CARA), which formulates resolution selection as an information-allocation problem. To evaluate the effectiveness of the proposed method, we conduct experiments on Kodak Lossless True Color Image Suite \cite{franzen_kodak_2013},\footnote{We refer to this as the ``Kodak dataset'' hereafter.} three gigapixel natural images (Pluto, Tokyo \cite{martel2021acorn, muller2022instant}, and Girl \cite{muller2022instant}) 
and four ultra-high resolution raw whole-slide images\footnote{Pluto: \href{https://github.com/computational-imaging/ACORN/blob/main/data/pluto.jpg}{Download link.} Tokyo: \href{https://www.flickr.com/photo_download.gne?id=29314390837&secret=b39ae4876e&size=o&source=photoPageEngagement}{Download link.} Girl: \href{https://upload.wikimedia.org/wikipedia/commons/8/82/21_Gigapixel_Total_Renovation_of_Girl_with_a_Pearl_Earring-Digital_Profoundism-Demo.jpg}{Download link.}WSIs: \href{https://www.ebi.ac.uk/biostudies/bioimages/studies/S-BIAD3807}{Download link.}}. 
Our contributions can be summarized as follows:
\begin{itemize}
\item We formulate resolution selection as an information-allocation problem to equalize the information encoded across layers, significantly improving parameter efficiency.
\item We propose an effective pixel-shuffle approach to mitigate information loss caused by hash collisions.
\item As elaborated in \cref{sec:wsi}, we curate, to the best of our knowledge, the first \textit{public} \textit{uncompressed} whole-slide image (WSI) dataset for academic research, providing a challenging benchmark for high-resolution image fitting. 

\item Extensive experiments demonstrate that CARA outperforms state-of-the-art baselines by a large margin; for instance, achieving equivalent fidelity to previous state-of-the-art using only 27.76\% of the parameters.
\end{itemize}

\section{Related Works}

\label{sec:relatedworks}
Implicit neural representations (INRs) parameterize signals as continuous functions through deep neural networks \cite{sitzmann2020implicit, tancik2020fourier}. In its standard formulation, an INR takes a coordinate as input and produces the signal value at that location \cite{sun2024recent}. 
However, conventional MLPs exhibit a spectral bias toward learning low-frequency components first, which can hinder the efficient recovery of high-frequency details \cite{sitzmann2020implicit}.
A large body of work mitigates this bias by proposing novel input encodings and activations, including Fourier feature mappings and periodic activations \cite{liu2024finer, sitzmann2020implicit, ramasinghe2022beyond, saragadam2023wire}, modifying the input coordinate representation \cite{tancik2020fourier, wang2021spline, landgraf2022pins}, or increasing model capacity \cite{hornik1991approximation,  liu2025kan}. These approaches expand the representable frequency content and improve optimization for fine-scale structures. Complementary to architectural changes, data-side transformations have been explored to improve INR fitting. Specifically, symmetric power transformations provide a reversible way to enforce range and approximate symmetry, demonstrating that input distributions materially affect INR optimization \cite{zhang2025enhancing}, and surprisingly, random pixel permutation \cite{10656124} can accelerate the convergence.

Another line of work attempts to improve rendering quality by introducing learnable feature grids queried via interpolation, followed by a lightweight decoder \cite{liu2020neural, muller2022instant}. Overall, these approaches query points from a discrete grid and use the interpolated features as the input for a decoder. Various data structures have been proposed to improve efficiency, including 2D feature planes \cite{reiser2023merf}, sparse octree \cite{yang2023tinc, yu2021plenoctrees}, and hyperplanes \cite{chan2022efficient, fridovich2023k, shue20233d, shi2024improved}. Such grid-based or hybrid representations substantially accelerate optimization and rendering. A particularly successful approach uses multiresolution hash-grid encoding \cite{muller2022instant}, which stores features in hash tables across a geometric progression of resolutions. This significantly accelerates inference because the bulk of the signal information is stored in the grid, which can be accessed via fast $O(1)$ lookups. 

Despite these advancements, most methods predominantly utilize data-agnostic data structures and necessitate manual parameter tuning to maximize reconstruction fidelity. Prior work typically relies on heuristic choices for the number of levels and geometric growth rates \cite{dai2026characterizing, muller2022instant}.

This study investigates grid resolution selection and collision modeling for multiresolution hash encoding. Instead of applying a fixed geometric progression of resolutions, collision-induced information degradation is quantified at each resolution, and grid resolutions are selected to equalize the information encoded per level. This approach prevents excessive capacity allocation to frequency bands with limited information, thereby enhancing parameter efficiency and improving reconstruction fidelity.

\section{Notation and Preliminaries}
\label{sec:NotationandPreliminaries}

\subsection{Notation}
\label{sec:Notation}

Scalars are denoted by non-bold letters (e.g., $a,\beta$), vectors by bold lowercase letters (e.g., $\mathbf{x}$), and tensors by bold uppercase letters (e.g., $\mathbf{Y}$). For a positive integer $n$, we write $[n]\triangleq\{1,\dots,n\}$. We use $\|\cdot\|$ for the Euclidean norm.

For a target image $\mathbf{Y}$ of spatial size $H\times W$, $\mathbf{Y}[i,j]$ denotes the entry at pixel index $(i,j)$. We associate pixel $(i,j)$ with its normalized center coordinate
\begin{equation}
\mathbf{x}_{ij}\triangleq\Big(\frac{j-\tfrac{1}{2}}{W},\,\frac{i-\tfrac{1}{2}}{H}\Big)\in[0,1]^2,
\label{eq:pixel_coord}
\end{equation}
so that all spatial frequencies are measured in cycles per pixel. 
%
%
Given an image tensor $\mathbf{Y}$ with intensity alphabet $\mathcal{V}\triangleq\{0,1,\dots,255\}$, we define its empirical entropy as
\begin{align}
H(\mathbf{Y})
= -\sum_{v\in\mathcal{V}} \hat{p}(v)\log_2 \hat{p}(v),
\label{eq:entropy}
\end{align}
where $\hat{p}(v)$ is the empirical probability mass function given by normalized histogram counts of $v$.
For a multi-channel image, we compute entropy per channel and sum the results.
For real-valued (e.g., filtered) signals, we apply a uniform quantizer before histogram estimation and compute $H(\cdot)$ on the quantized samples.
Specifically, intensity-valued images use an 8-bit quantizer $Q_{8}(\cdot)$ with alphabet $\mathcal{V}$, whereas residual signals use a signed 9-bit quantizer $Q_{\Delta}(\cdot)$ with alphabet $\mathcal{V}_{\Delta}\triangleq\{-255,-254,\ldots,255\}$.
Unless stated otherwise, $H(\cdot)$ denotes the empirical entropy computed over the corresponding fixed discrete alphabet.

\subsection{Instant Neural Graphics Primitives}
\label{sec:prelim_inr_ingp}
Implicit neural representations (INRs) model a discrete signal as samples of a continuous function 
$\mathfrak{f}_\theta:\Omega \rightarrow \mathbb{R}^{c}$, where $c$ denotes the number of output channels and $\mathbf{x} \in \Omega \subset \mathbb{R}^{2}$ represents spatial coordinates in the input domain.  Given a set of training pairs $\{(\mathbf{x}_m, \mathbf{y}_m)\}_{m=1}^{M}$, the network parameters $\theta$ are optimized by minimizing a reconstruction loss,
\begin{equation}
\min_{\theta} \; \frac{1}{M} \sum_{m=1}^{M} 
\left\| \mathfrak{f}_\theta(\mathbf{x}_m) - \mathbf{y}_m \right\|^2.
\label{eq:inr_objective}
\end{equation}

\begin{figure}[tb]
\centering
\includegraphics[width=\linewidth]{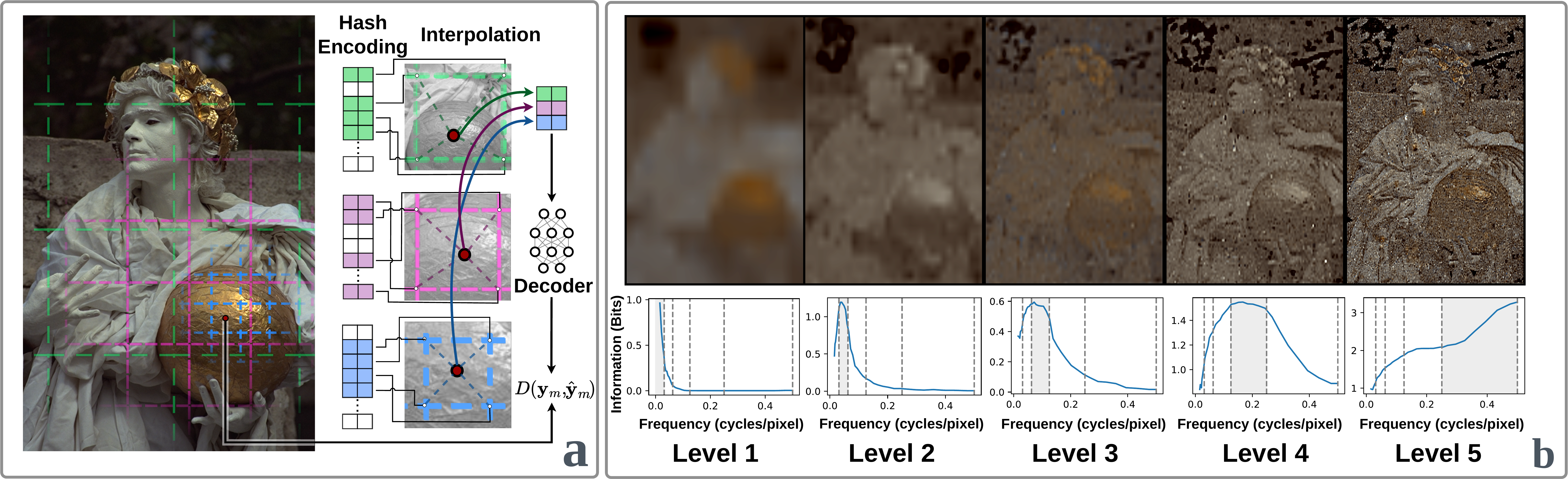}
\caption{Panel (a) illustrates that Instant-NGP factorizes an implicit neural representation into a multiresolution hash encoding and a lightweight MLP decoder. A query coordinate is interpolated within each grid level. The resulting per-level features are concatenated. The decoder is trained by minimizing the reconstruction distortion $D(\mathbf{y}_m, \hat{\mathbf{y}}_m)$. Panel (b) shows that after training an Instant-NGP model with a 5-level hash grid, we isolate the contribution of each level by retaining only the target level. All other levels’ hash-table entries are replaced with their mean feature value. The resulting reconstructions demonstrate clear spectral stratification. Lower-resolution levels recover smooth, low-frequency structures. Higher-resolution levels progressively contribute fine, high-frequency components. This behavior is quantified by the information density $\mathcal{I}_f$, defined in \Cref{eq:informband}, and plotted against frequency. Vertical dashed lines indicate the sampling frequency of each feature grid level. This confirms that different hash-grid levels encode information from different frequency bands.}
\label{fig:INGP_info}
\end{figure}
Coordinate-based MLPs can represent complex signals, but their optimization often suffers from spectral bias \cite{mildenhall2021nerf}, exhibiting a preference for low-frequency components over high-frequency ones. This can slow convergence and reduce fidelity when the target contains rich high-frequency content \cite{tancik2020fourier,NEURIPS2020_53c04118}.

Instant neural graphics primitives (Instant-NGP) alleviates this limitation by combining a lightweight decoder with a trainable multiresolution hash encoding that provides spatially localized, multi-scale features \cite{muller2022instant}. Specifically, Instant-NGP parameterizes an INR as
\begin{equation}
\mathfrak{f}_\theta(\mathbf{x})=\mathfrak{g}_\theta\left(\phi(\mathbf{x})\right), 
\quad
\phi(\mathbf{x})=\big[\phi_1(\mathbf{x});\dots;\phi_L(\mathbf{x})\big],
\label{eq:ingp_factorization}
\end{equation}
where $\mathfrak{g}_\theta$ is a lightweight MLP and $\phi(\mathbf{x})$ concatenates the features from $L$ resolution levels. 


Assuming $\mathbf{x}$ is normalized to $\Omega=[0,1]^2$, level $\ell$ defines a $2$D grid with resolution $N_\ell$ per axis. Vertex embeddings are stored in a trainable table $\mathbf{E}_\ell \in \mathbb{R}^{T_\ell \times F}$, where $T_\ell$ is the table size and $F$ is the feature dimension. A hash function $\mathfrak{h}_\ell:\mathbb{Z}^2\rightarrow [T_\ell]$ maps integer grid coordinates to table indices, enabling memory-efficient parameterization at high effective resolutions.

For a query $\mathbf{x}$, the level feature $\phi_\ell(\mathbf{x})\in\mathbb{R}^F$ is obtained by bilinear interpolation of the embeddings at the four neighboring grid vertices:
\begin{equation}
\phi_\ell(\mathbf{x})
= \sum_{\boldsymbol{\delta}\in\{0,1\}^2}
w_{\ell,\boldsymbol{\delta}}(\mathbf{x})\;
\mathbf{E}_\ell\!\left[
\mathfrak{h}_\ell\!\left(\big\lfloor N_\ell\mathbf{x}\big\rfloor+\boldsymbol{\delta}\right)
\right],
\label{eq:hashgrid_level}
\end{equation}
where $w_{\ell,\boldsymbol{\delta}}(\mathbf{x})$ are the standard bilinear interpolation weights that sum to $1$. Instant-NGP chooses the per-level resolutions using a data-agnostic geometric progression,
$N_\ell=\lfloor N_{\min}\, b^{\ell-1}\rfloor$ for $\ell\in\{1,\dots,L\}$ and a predefined base $b>1$. This yields grid spacing $\Delta_\ell \propto 1/N_\ell$, so higher levels (larger $N_\ell$) can represent finer spatial variation. Concatenating features across levels provides a multi-scale representation: low-resolution levels primarily capture coarse, low-frequency structure, while high-resolution levels encode finer, high-frequency details. As illustrated in \Cref{fig:INGP_info} panel (b), decoding with individual levels shows that higher-resolution grids recover high-frequency components, whereas lower-resolution grids represent the coarse structure of the signal.

\begin{figure}[tb]
  \centering
  \includegraphics[width=\linewidth]{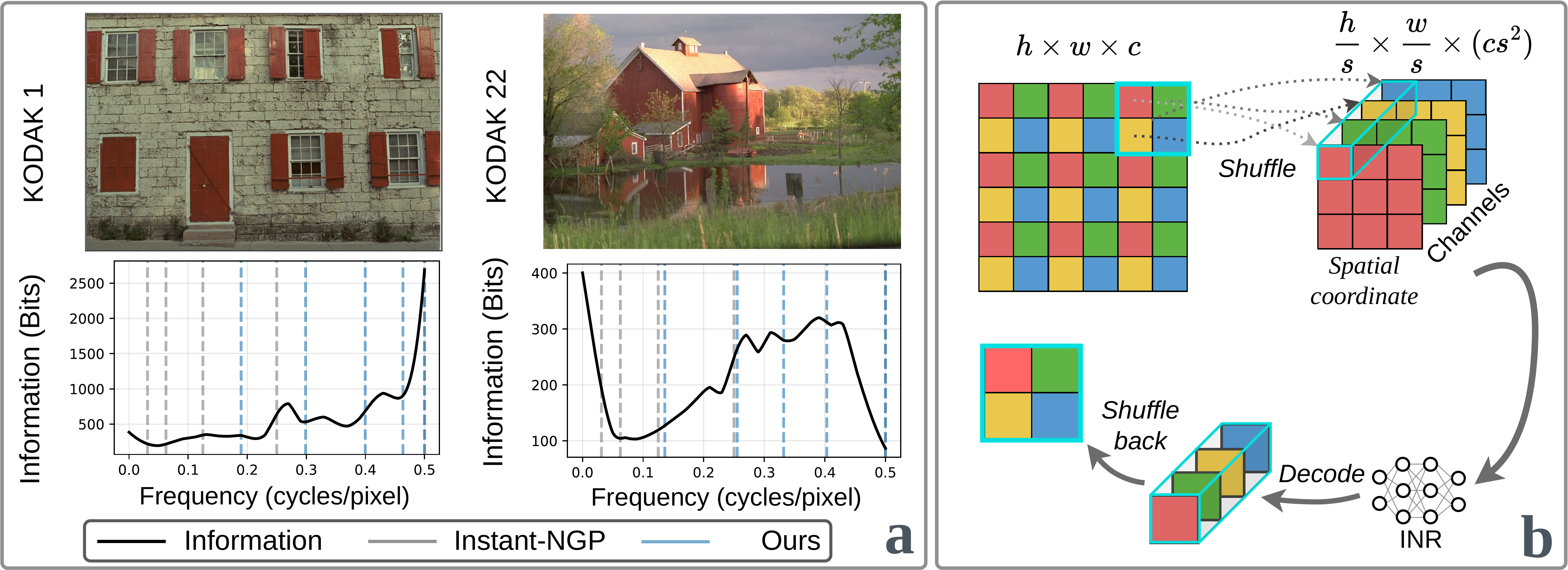}
  \caption{Method overview. Panel (a): different images have different information density $\mathfrak{I}(f)$, defined in \Cref{eq:informband}, for two Kodak images, showing that different images concentrate information at different frequency ranges; the vertical dashed lines indicate the frequency cutoffs implied by the fixed Instant-NGP schedule (gray) versus our collision-aware resolution schedule (blue), which aims to equalize the information being encoded by each level. Panel (b): pixel shuffling to mitigate collision-induced information loss. We partition the image into non-overlapping patches, then shuffle them into channels, thereby reducing the hash collision.
  }
  \label{fig:Method}
\end{figure}

\section{Methodology}
\label{sec:methodology}
As discussed previously, the multiresolution hash encoding in Instant-NGP can be interpreted as a sampling-and-interpolation system \cite{chen2024far}. 
At level $\ell$, a grid with resolution $N_\ell$ per axis induces an effective sampling interval of 
$\Delta_\ell^{(x)} = w/N_\ell$ and $\Delta_\ell^{(y)} = h/N_\ell$ pixels along the horizontal and vertical directions, respectively. 
Under this interpretation, the corresponding Nyquist frequencies (in cycles per pixel) are
\begin{equation}
f_{\text{Nyq},\ell}^{(x)} = \frac{N_\ell}{2w}, 
\qquad 
f_{\text{Nyq},\ell}^{(y)} = \frac{N_\ell}{2h}.
\end{equation}
Frequencies beyond these limits cannot be represented without aliasing at level $\ell$ and must therefore be captured by finer-resolution levels or by the MLP decoder.

This observation suggests that each level $\ell$ primarily contributes to modeling a specific spatial frequency band determined by its resolution. 
Since frequencies are measured in cycles per pixel, a level with resolution $N_\ell$ per axis has an effective Nyquist limit $f_{\text{Nyq},\ell}^{(x)} = N_\ell/(2w)$ and $f_{\text{Nyq},\ell}^{(y)} = N_\ell/(2h)$. 
Hence, there exists an approximate correspondence between a cutoff frequency $f$ and a grid resolution $N_f = 2wf$ (assuming square images for simplicity).

To quantify the information content associated with a frequency band centered at $f$, we measure the entropy \cite{yenormalized,ICLR2024_718a3c5c}increment between two adjacent spectral cutoffs.

Let $\mathbf{Y}_{\le f}$ denote the low-pass filtered image retaining frequencies up to $f$ (in cycles per pixel). 
We define the band information density as
\begin{equation}
\mathfrak{I}(f) 
= 
H\!\left(\mathbf{Y}_{\le f+\Delta f} - \mathbf{Y}_{\le f}\right)
\cdot N_f^{2},
\label{eq:informband}
\end{equation}
where $H(\cdot)$ denotes the empirical Shannon entropy computed over pixel intensities, 
$\Delta f$ is a small frequency increment (in cycles per pixel), and $N_f$ is the grid resolution corresponding to cutoff frequency $f$. 
The multiplicative factor $N_f^{2}$ is the number of spatial samples at that resolution.
\begin{remark}[Why entropy as an information proxy]
Shannon's source coding theorem \cite{shannon1948mathematical, 10900607,10619241, ye2026asmil} implies that entropy quantifies the amount of information carried by a source. 
Accordingly, the quantity in \Cref{eq:informband} can be interpreted as the new information introduced when increasing the effective cutoff from $f$ to $f+\Delta f$ (measured in cycles per pixel). 
The increment $H\!\left(\mathbf{Y}_{\le f+\Delta f}-\mathbf{Y}_{\le f}\right)$ captures the average additional uncertainty (bits per sample) contributed by this narrow frequency band, while $N_f^{2}$ denotes the number of spatial samples at the corresponding grid resolution. 
Their product therefore provides an estimate of the total information content of that refinement step. In this sense, $\mathfrak{I}(f)$ serves as a practical proxy for the information budget that a grid level should allocate in order to faithfully model structures up to cutoff frequency $f$. 
We ablate alternative information proxies in \Cref{sec:ablation_info_proxy}.
\end{remark}
Equipped with the information density measure $\mathfrak{I}(f)$, we visualize its distribution for two representative images from the Kodak dataset \cite{franzen_kodak_2013} in Panel (a) of \Cref{fig:Method}. 
The two images exhibit markedly different spectral information distributions. 
Kodak~1 shows relatively limited low-frequency information, with a substantial portion of its entropy concentrated in higher-frequency bands. 
In contrast, Kodak~22 distributes its information more heavily across low- and mid-frequency regions, with comparatively reduced high-frequency content.

Despite such variability, Instant-NGP employs a data-agnostic geometric resolution schedule designed to uniformly cover a wide frequency range. 
In the next section, we introduce our method for selecting per-level resolutions adaptively based on the measured information distribution of the input.

\subsection{Collision-Aware Resolution Adaptation}
\label{Sec:method-CARA}
In this section, we present \emph{collision-aware resolution adaptation (CARA)}, which assigns per-level grid resolutions according to the input's measured information density. 
Our goal is to determine a resolution schedule $\{N_\ell\}_{\ell=1}^{L}$ such that each level is allocated a comparable effective information budget.

Let $\mathbf{Y}_{\le f}$ denote the input signal low-pass filtered to cutoff frequency $f$ (in cycles per pixel).\footnote{We study the impact of different low-pass filters in the supplementary material.}
Building upon the information density measure $\mathfrak{I}(f)$ defined in \Cref{eq:informband}, we first aim to equalize the information allocated to each hash-grid level. To this end, we partition the frequency axis into $L$ disjoint bands $\{\mathcal{F}_\ell\}_{\ell=1}^L$ such that each band carries the same total information:
\begin{align}
\mathcal{I}_\ell \triangleq \sum_{f \in \mathcal{F}_\ell} \mathfrak{I}(f)  = \tau, \quad \forall \ell \in [L],
\end{align}
where $\tau$ is a shared target information budget. 
However, in hash encoding, the number of distinct trainable parameters at level $\ell$ is constrained by the hash table size $T$. 
Let $n_\ell = N_\ell^2$ denote the number of distinct grid vertices at that level. 
Since $n_\ell$ may exceed $T$, hashing induces collisions, where multiple grid vertices map to the same table entry. 
Assuming uniform hashing, the following proposition characterizes the expected reduction in usable capacity due to collisions. \\

\begin{proposition}[Effective Capacity under Uniform Hash Collisions]
\label{prop:hash_efficiency}
Suppose $n_\ell$ keys are inserted into a hash table of size $T$, with load factor $\alpha_\ell = n_\ell/T$. 
Under uniform hashing,
the expected number of distinct entries per inserted key is 
\begin{equation}
w_\ell^{\mathrm{hash}} = \frac{1 - e^{-\alpha_\ell}}{\alpha_\ell}. \label{eq:whash}
\end{equation}
Accordingly, if $B$ bits are stored across the $n_\ell$ keys, the expected retrievable information scales as
\begin{equation}
B_{\mathrm{eff}} = B \cdot w_\ell^{\mathrm{hash}}.
\end{equation}
\end{proposition}

We defer the proof to the supplementary material due to the space limit. Following \Cref{prop:hash_efficiency}, we incorporate the collision factor into the per-level information by defining the effective information at level $\ell$ as
\begin{equation}
    \mathcal{I}_{\mathrm{eff},\ell}
    =
    \mathcal{I}_{\ell}
    \cdot
    w^{\mathrm{hash}}_\ell.
\end{equation}
After obtaining the collision-aware effective information, we determine the per-level resolutions so that different frequency bands carry comparable amounts of effective information.

Rather than discretizing frequency directly, we consider a monotone candidate set of grid resolutions 
$\{N^{(k)}\}_{k=1}^{K}$ with $N^{(1)} < \cdots < N^{(K)}$, 
each implying a Nyquist cutoff in cycles per pixel. 
Let $Y^{(k)}$ denote the input low-pass filtered to the Nyquist limit corresponding to resolution $N^{(k)}$, and define 
$H_k \triangleq H(Y^{(k)})$, with $H_0$ a coarse reference.

For each candidate $k$, we define the associated collision efficiency
\begin{equation}
w^{\mathrm{hash}}_k 
\triangleq 
\frac{1-e^{-\alpha_k}}{\alpha_k}, \quad
 \alpha_k \triangleq \frac{(N^{(k)})^{2}}{T}.
\end{equation}
A resolution schedule is specified by indices 
\begin{equation}
0 = k_0 < k_1 < \cdots < k_L = K,
\end{equation}
which partition the spectrum into $L$ adjacent bands. 
The effective information carried by band $\ell$, corresponding to the interval between cutoffs $k_{\ell-1}$ and $k_\ell$, is defined as
\begin{equation}
\mathcal{I}_\ell
\triangleq w^{\mathrm{hash}}_{k_\ell}\,
\sum_{f \in \mathcal{F}_\ell} \mathfrak{I}(f),
\label{eq:band_info}
\end{equation}
which explicitly accounts for collision-induced capacity reduction at the selected resolution. We then select the cutoffs by solving the min--max optimization
\begin{equation}
\min_{0<k_1<\cdots<k_{L-1}<K} \quad
\max_{\ell\in[L]}
\mathcal{I}_\ell,
\label{eq:minmax}
\end{equation}
which seeks the tightest achievable uniform upper bound on per-band effective information.
\begin{remark}[Why balance effective information and why min--max]
Each hash-grid level has a bounded effective capacity: it uses a fixed-size table and its usable information is further reduced by collisions. Meanwhile, levels specialize to different frequency ranges (\Cref{fig:INGP_info} (b)), so reconstruction quality is often constrained by the most overloaded band: if one band contains much more information than its level can represent, the residual must be absorbed by other levels or the lightweight decoder, creating a capacity bottleneck, while over-allocating to a low-information band wastes parameters. We therefore balance per-band effective information as a load-balancing principle under a fixed per-level budget. Exact equality $\mathcal{I}_\ell=\tau$ is generally infeasible under discrete candidate cutoffs, so we minimize $\max_{\ell\in[L]} \mathcal{I}_\ell$ \cite{boyd2004convex}, \ie, the tightest achievable uniform upper bound on per-band load, preventing any single band from becoming capacity-limiting.
\end{remark}
The optimization problem in \Cref{eq:minmax} admits an exact solution via dynamic programming, as the objective decomposes over ordered band partitions with optimal substructure. 
We defer the detailed algorithm and pseudocode to the supplementary material. 
The resulting computational complexity is $O(LK^2)$, which is modest for typical candidate sizes $K$ and negligible compared to the overall training cost of the INR, e.g., adding only 0.7 seconds to pre-processing for the Pluto gigapixel image.

While CARA optimizes the allocation of grid resolution across frequency bands, the effective capacity of each level remains constrained by hash collisions. 
To further alleviate collision-induced information loss without increasing the hash table size, we introduce an efficient strategy dubbed pixel shuffling.

\subsection{Reducing Hash Collisions via Pixel Shuffling}
\label{sec:MethodPixelShuffleing}
The performance of Instant-NGP improves as the hash table size $T$ increases, since a larger table reduces the load factor and thereby improves the collision efficiency $w^{\mathrm{hash}}$. 
However, enlarging $T$ incurs a linear increase in both memory usage and the number of trainable parameters. To reduce collisions without increasing $T$, we apply the pixel shuffle operation \cite{7780576, chi2025plug}. 
Given a target image $\mathbf{Y} \in \mathbb{R}^{h \times w \times c}$ and an integer shuffle factor $s$, the pixel shuffle transform $\mathcal{S}_s$ rearranges $\mathbf{Y}$ into a lower-resolution, higher-channel representation
\begin{equation}
\tilde{\mathbf{Y}} = \mathcal{S}_s(\mathbf{Y})
\in 
\mathbb{R}^{\frac{h}{s} \times \frac{w}{s} \times (c s^2)}.
\end{equation}
Each spatial coordinate in $\tilde{\mathbf{Y}}$ aggregates an $s \times s$ patch from the original image. 
Accordingly, the INR predicts a higher-dimensional vector 
\begin{equation}
\hat{\tilde{\mathbf{y}}}_m = \mathfrak{f}_\theta(\tilde{\mathbf{x}}_m) \in \mathbb{R}^{c s^2},
\end{equation}
and the full-resolution reconstruction is recovered via the inverse transform 
\begin{equation}
\hat{\mathbf{Y}} = \mathcal{S}_s^{-1}(\hat{\tilde{\mathbf{Y}}}).
\end{equation}

Importantly, pixel shuffling preserves all original information while reducing the spatial resolution of the coordinate domain by a factor of $s$ along each axis. 
Consequently, at level $\ell$, the number of distinct grid vertices decreases from $N_\ell^2$ to $(N_\ell/s)^2$. 
Under a fixed hash table size $T$, the load factor becomes
\begin{equation}
\alpha_\ell' 
= \frac{(N_\ell/s)^2}{T}
= \frac{\alpha_\ell}{s^2}.
\end{equation}
By \Cref{prop:hash_efficiency}, reducing the load factor increases the collision efficiency factor $w^{\mathrm{hash}}$, particularly when the original $\alpha_\ell$ is large. Thus, pixel shuffling effectively improves usable capacity and rendering fidelity without increasing $T$. We defer the ablation study on $s$ to the supplementary material.

\begin{figure}[b]
  \centering
  \includegraphics[width=\linewidth]{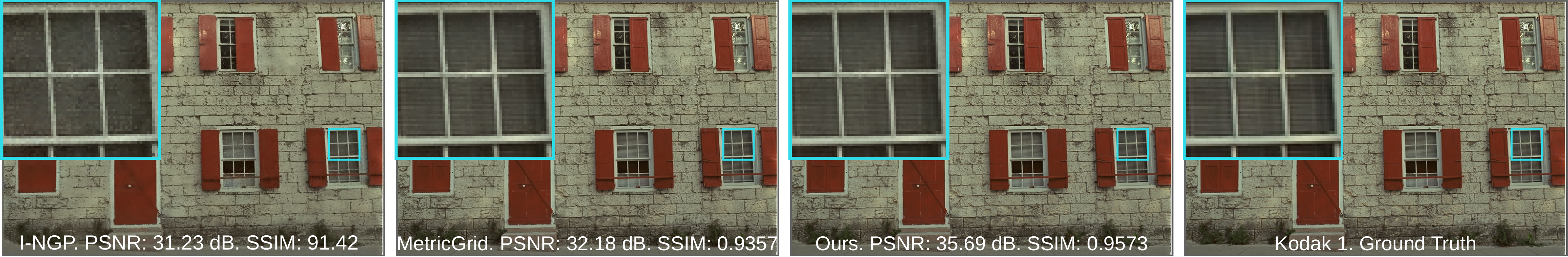}
  \caption{Qualitative comparison on the Kodak dataset. By selecting per-level resolutions (rather than using a fixed, data-agnostic schedule), our method better preserves high-frequency structures and textures, yielding sharper reconstructions than I-NGP and MetricGrid. (Best viewed digitally with zoom).}
  \label{fig:KodakVis}
\end{figure}

\section{Experiments}
\label{sec:Experiments}
We evaluate the proposed method on three benchmarks: Kodak dataset, three gigapixel natural images, and ultra-high-resolution whole-slide images (WSIs). Unless otherwise specified, all models are trained with the Adam optimizer \cite{kingma2014adam} with $(\beta_1,\beta_2,\epsilon)=(0.9,0.99,10^{-15})$, cosine annealing learning-rate decay, and an $\ell_2$ reconstruction loss. Across all experiments, we employ the same multiresolution hash-grid architecture with feature dimension fixed to 2 at every level and the same number of levels for all methods and parameter budgets. To match different parameter budgets while keeping the architecture depth unchanged, we vary the hash-table size. For all experiments, we use anti-aliased resizing as the low-pass filter. For Kodak, we set  $K = 30$ with uniform spacing in the frequency domain; for gigapixel images and WSIs, we set $K=200$. Additional implementation details and wall-clock cost are provided in the supplementary material.
\begin{table}[tb]
\caption{Image fitting results on the Kodak dataset. All methods are trained with the same number of iterations and a comparable number of parameters. Our method achieves the highest PSNR and SSIM while requiring fewer parameters compared to all other methods. \textbf{Bold} and \underline{underlined} values denote the best and second-best results.}
\label{tab:KodakResults}
  \centering
  \resizebox{\textwidth}{!}
{\begin{tabular}{ccccccccc}
 \toprule
\rowcolor{mygray} Method     & WIRE \cite{saragadam2023wire}  & SIREN \cite{NEURIPS2020_53c04118}   & SCONE \cite{li2024learning}   & I-NGP \cite{muller2022instant} & NFFB  \cite{wu2023neural} & NeuRBF \cite{chen2023neurbf}& MetricGrid \cite{wang2025metricgrids}& Ours   \\ 
Rep.       & Implicit & Implicit & Implicit & Hybrid & Hybrid & Hybrid & Hybrid     & Hybrid \\ \hline
 Params (K) & 373      & 207      & 207      & 206    & 208    & 207    & 207        & 205 \\
\textbf{PSNR} $\uparrow$      & 37.59    & 38.70    & 39.72    & 37.06  & 38.77  & 38.70  & \underline{39.73}     & \textbf{40.03}\\
  \textbf{SSIM}  $\uparrow$     & 0.9596   & 0.9512   & \underline{0.9662}   & 0.9386 & 0.9537 & 0.9488 & 0.9568     & \textbf{0.9714}   \\ \bottomrule
\end{tabular}}
\end{table}
\begin{table}[tb]
\centering
\caption{Quantitative results of image fitting on the Kodak dataset. Comparing our method against the Gaussian primitive-based approach. \textbf{Bold} and \underline{underlined} values denote the best and second-best results, respectively.}
\label{Tab:ScalingKodak}
\resizebox{\textwidth}{!}{\begin{tabular}{ccccccccc}
\toprule
\rowcolor{mygray} Method     & I-NGP \cite{muller2022instant} & NeuRBF \cite{chen2023neurbf} & Gaussian Splatting  \cite{kerbl20233d}    & GaussianImage \cite{zhang2024gaussianimage} & \multicolumn{2}{c}{MetricGrid  \cite{wang2025metricgrids} } & \multicolumn{2}{c}{Ours}   \\
Rep.       & Hybrid & Hybrid & Explicit & Explicit      & \multicolumn{2}{c}{Hybrid}     & \multicolumn{2}{c}{Hybrid} \\
\cmidrule(lr){2-2} \cmidrule(lr){3-3} \cmidrule(lr){4-4} \cmidrule(lr){5-5} \cmidrule(lr){6-7} \cmidrule(lr){8-9} 
Params (K) & 300    & 337    & 3540     & 560           & 335            & 533           & 341          & 471         \\
\textbf{PSNR} $\uparrow$       & 43.88  & 43.78  & 43.69    & 44.08         & 44.51          & 45.77         & \underline{46.24}        & \textbf{48.00}       \\
\textbf{SSIM} $\uparrow$      & 0.9976 & 0.9964 & \textbf{0.9991}   & 0.9985        & ~0.9778         & ~0.9897~        & ~0.9906~       & \underline{0.9987} ~     \\ \hline
\end{tabular}}
\end{table}

\subsection{Kodak Dataset}

We first evaluate our approach on the Kodak dataset \cite{franzen_kodak_2013}, which consists of 24 natural images with resolution of $768\times 512$. 
We report PSNR and SSIM \cite{1284395} to assess reconstruction fidelity. We compare against existing methods under comparable parameter budgets (around 207K), and train all models for the same number of iterations. Due to their relatively low resolution, we do not apply pixel shuffle to the Kodak dataset. The results are presented in \Cref{tab:KodakResults}. As seen, our method achieves the best image fidelity in terms of both PSNR and SSIM with fewer parameters.

We further compare our method against Gaussian primitive-based representations, including Gaussian Splatting (GS) \cite{kerbl20233d} and GaussianImage \cite{zhang2024gaussianimage}. To align the comparison across parameter regimes, we report CARA under two budgets in \Cref{Tab:ScalingKodak}. At 471K parameters, CARA uses only $13\%$ of the trainable parameters of GS (3540K) while achieving substantially higher PSNR. Moreover, CARA attains the best PSNR overall and improves upon the strongest hybrid baseline, MetricGrid \cite{wang2025metricgrids}, by $2.23$~dB, highlighting the benefit of assigning resolutions based on the input image.
\begin{figure}[t]
\centering
\includegraphics[width=\linewidth]{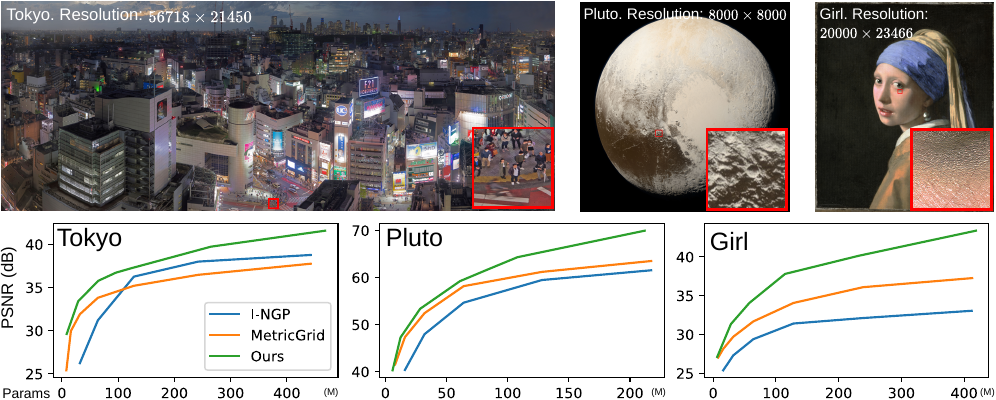}
\caption{Gigapixel Image Fitting Performance. \textbf{Top:} Gigapixel images (Tokyo, Pluto, and Girl) with magnified crops highlighting fine-grained details. (Best viewed digitally with zoom). \textbf{Bottom:} Quantitative comparison of reconstruction quality (PSNR) vs. the number of trainable parameters (M). Across all three images, our method consistently outperforms Instant-NGP (I-NGP) and MetricGrid, achieving a superior Pareto frontier and higher fidelity at every parameter scale.}
\label{Fig:ResultsGigaTradoff}
\end{figure}
\subsection{Gigapixel Natural Images}
To evaluate CARA's performance at scale, we conducted experiments on three canonical gigapixel images: \emph{Pluto}, \emph{Tokyo}, and \emph{Girl With a Pearl Earring} (Girl). These images pose a significant challenge for implicit neural representations, given their high resolution and the need to preserve fine-grained details across vast spatial domains. We compare the reconstructed image fidelity under multiple parameter budgets against Instant-NGP \cite{muller2022instant}, NeuRBF \cite{chen2023neurbf}, and MetricGrid \cite{zhang2026widget2code}. \Cref{Fig:ResultsGigaTradoff} reports PSNR versus the number of trainable parameters, where our method consistently achieves a better fidelity–parameter trade-off across all three images. Quantitatively, \Cref{tab:GigaResults} shows that we match MetricGrid’s best reported fidelity ($37.26~dB$) using only $114.61M$ parameters, $27.76\%$ of MetricGrid's $412.8M$ parameter setting, demonstrating substantially improved parameter efficiency in the high-fidelity regime. Moreover, at a comparable large-model parameter budget, our method improves PSNR by $6.11~dB$ over the strongest baseline, demonstrating that collision-aware, data-adaptive resolution allocation yields consistent fidelity gains even in the high-capacity regime. Notably, in the low-parameter regime our method remains competitive, achieving higher image quality with $6.31M$ trainable parameters, which is $21.3\%$ fewer than MetricGrid's $8.02M$ setting. 
\begin{table}[t]
\centering
\caption{Numerical comparison on gigapixel girl image fitting. We report PSNR (dB), SSIM and the corresponding number of trainable parameters (in millions). Our method maintains competitive fidelity in both the low-parameter and high-parameter regimes.}
\label{tab:GigaResults}
\resizebox{\textwidth}{!}{
\begin{tabular}{cccccccccccc}
\toprule
\rowcolor{mygray} Method     & \multicolumn{3}{c}{~~I-NGP \cite{muller2022instant}~~} & ~~NeuRBF \cite{chen2023neurbf}~~  & \multicolumn{3}{c}{~~MetricGrid \cite{wang2025metricgrids}~~} & \multicolumn{4}{c}{Ours}       \\ 
\cmidrule(lr){2-4} \cmidrule(lr){5-5}\cmidrule(lr){6-8}\cmidrule(lr){9-12} 
Params (M) & 8.01 & 32.01 & 412.8 & 115.4 & 8.02     & 32.02    & 412.8~    & ~8.02~  & ~28.24~ & ~114.6~ & 418.5 \\
\textbf{PSNR} $\uparrow$      & 21.51 & 27.3 & 33.05 & 31.56  & 27.03    & ~29.70 ~    & ~37.06~    & ~27.53~ & 31.33~ & 37.07~  & 43.37 \\ 
\textbf{SSIM} $\uparrow$      & ~ 0.579 ~ & ~ 0.896 ~ & ~0.959~ & 0.960  & ~0.817~  & ~0.957 ~   & ~0.994~    & ~0.897~ & ~0.984~ & 0.998~  & 0.999 \\ \hline
\end{tabular}}
\end{table}
\begin{figure}[tb]
\centering
\includegraphics[width=\linewidth]{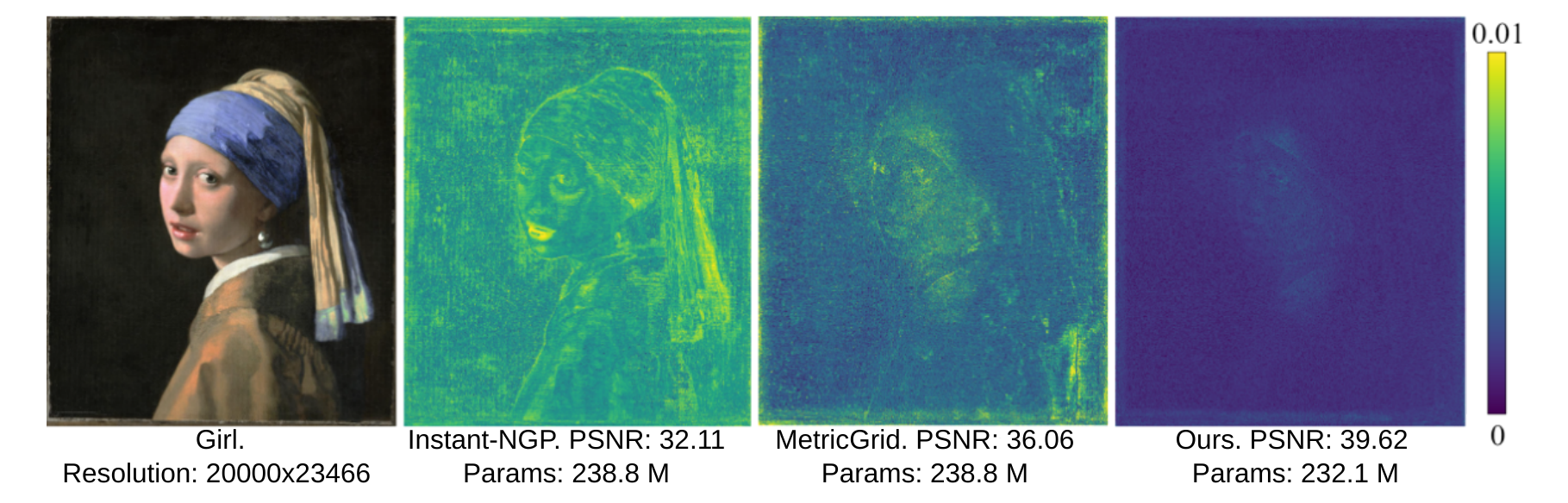}
\caption{Qualitative comparison on a gigapixel image. The leftmost presents the ground-truth image, while the three panels on the right visualize the per-pixel $\ell_2$ error maps of I-NGP, MetricGrid, and the proposed method, respectively.}
\label{Fig:GirlResidual}
\end{figure}
Finally, \Cref{Fig:GirlResidual} visualizes per-pixel $\ell_2$ error maps on a gigapixel example; under matched training protocols, our approach yields visibly lower residual error than both baselines, corroborating the benefit of adaptive resolution allocation. Similar trends can be observed for Tokyo and Pluto; we defer the detailed analysis and visualization to the supplementary material due to space constraints. 
\subsection{Whole-Slide Images} \label{sec:wsi}
\begin{table}[t]
\centering
\caption{Numerical comparison on WSI image fitting. We report PSNR (dB) and the corresponding number of trainable parameters. Our method maintains competitive reconstruction fidelity in the low-parameter regime, while the high-capacity results are provided explicitly to facilitate direct numerical comparison in the large-model setting.}
\label{tab:WSIResults}
\resizebox{\textwidth}{!}{\begin{tabular}{cccccccccccc}
\toprule
\rowcolor{mygray}Method     & \multicolumn{3}{c}{~ I-NGP \cite{muller2022instant} ~} & ~ NeurRBF \cite{chen2023neurbf} ~& ~ CINR \cite{Lee_Convolutional_MICCAI2024} ~ &  \multicolumn{3}{c}{ ~ MetricGrid \cite{wang2025metricgrids} ~ } & \multicolumn{3}{c}{Ours} \\
\cmidrule(lr){2-4} \cmidrule(lr){5-5} \cmidrule(lr){6-6} \cmidrule(lr){7-9} \cmidrule(lr){10-12}
Params (M)~ & 95.45~ & 214.02 ~ &  402.6  & 161.48  & 157.24 & 95.46 ~  & 199.25 ~ &  390.01  & ~ 18.24       &~ 93.53 & ~ 233.57 \\
\textbf{PSNR} $\uparrow$ ~ & 40.97 ~ &  41.68 ~ &  42.06  & 40.30  & 40.69  & 41.06 ~ &  42.19 ~ & ~ 43.03 ~  & ~  41.24      &  ~42.24 & ~ 44.02 \\ \hline
\end{tabular}}
\end{table}
\begin{figure}[]
  \centering
  \includegraphics[width=\linewidth]{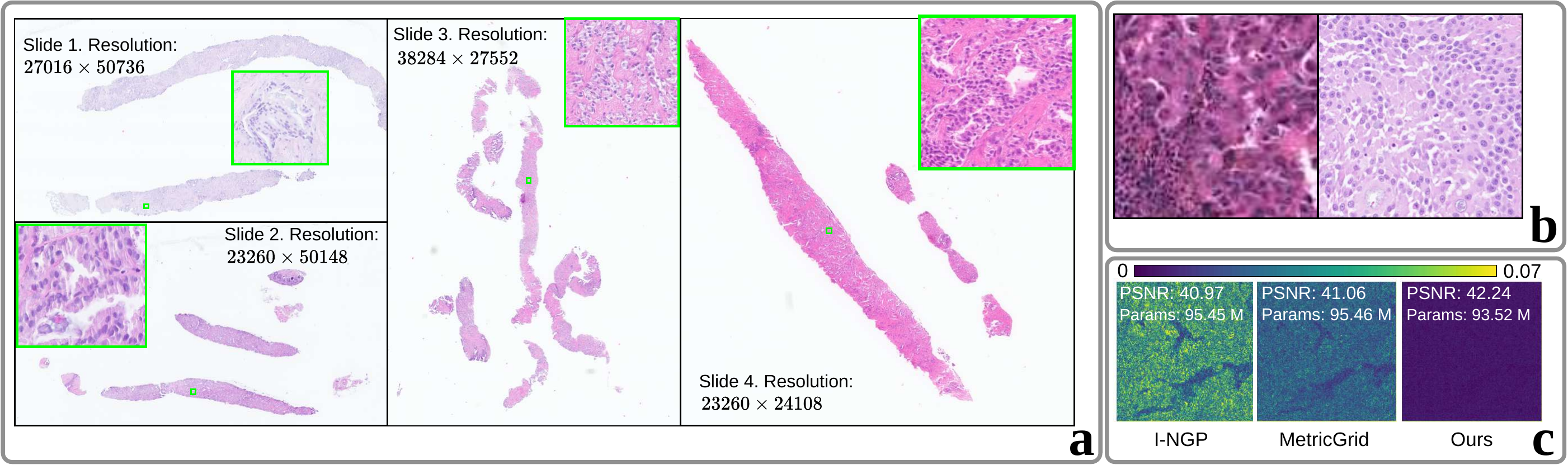}
  \caption{WSIs used for evaluation. Panel (a) four ultra-high-resolution WSIs from our collection; green boxes indicate regions shown at higher magnification, highlighting both global tissue architecture and cell-level morphology. Panel (b) patch-level visual comparison between a WSI patch from prior literature (left) and a patch from our raw scans (right): aggressive JPEG compression in the literature data introduces visible blocking artifacts, whereas our raw WSIs preserve fine structural detail. (Best viewed digitally with zoom). Panel (c) per-pixel $\ell_2$
 reconstruction error maps on a representative patch; CARA produces consistently lower error, indicating improved image fidelity.
  }
  \label{fig:WSI_results}
\end{figure}
WSIs \cite{hosseini2019atlas, omoush2026role} are ultra–high-resolution digital scans of histopathology slides, and constitute the primary data modality for computational pathology \cite{hosseini2024computational,ye2026asmil}. Unlike natural images, WSIs contain micro-scale cellular morphology and macro-scale tissue architecture, requiring models to accurately represent structures across a wide spectrum of spatial frequencies. Their extreme spatial resolution and dense fine-grained texture pose significant challenges for conventional discrete representations. 

In this section, we evaluate INRs to represent WSIs. Most publicly available WSI datasets are stored with aggressive JPEG compression \cite{wallace1991jpeg}, which introduces artifacts that can negatively affect downstream tasks \cite{ignatov2024nct}. To enable faithful evaluation, we collected four uncompressed $H\&E$ stained WSIs\footnote{All patient-identifiable information has been rigorously removed to ensure privacy protection. The downsampled WSIs and dataset details are available in the supplementary material. The full-resolution images will be released upon acceptance.} \cite{fischer2008hematoxylin} at $40\times$ magnification (\Cref{fig:WSI_results} Panel (a)). Compared with compressed WSIs used in prior work (\Cref{fig:WSI_results} Panel (b)), our raw scans preserve fine structural detail. We benchmark CARA against four INR baselines and summarize the results in \Cref{tab:WSIResults}. CARA consistently outperforms the state-of-the-art methods across all parameter settings.
\begin{table}[tb]
\centering
\caption{Ablation of the collision factor $w^{\text{hash}}$ on Kodak. Disabling collision modeling degrades PSNR/SSIM while typically requiring more parameters to reach comparable quality, highlighting the importance of explicitly accounting for hash collisions.}
\label{Tab:ablationcollision}
\resizebox{0.6\textwidth}{!}{\begin{tabular}{cccc|ccc}
\toprule
\rowcolor{mygray}  & \multicolumn{3}{c|}{W.O. Collision Factor} & \multicolumn{3}{c}{W. Collision Factor} \\
Params (K) & 374          & 403          & 484         & 205         & 341         & 471         \\
\textbf{PSNR}  $\uparrow$      & 40.32        & 43.97        & 47.97       & 41.03       & 46.24       & 48.00       \\
\textbf{SSIM}  $\uparrow$      & ~0.9674~ & ~0.9973~ & ~0.9979~ & ~0.9714~ & ~0.9906~ & ~0.9987~ \\ \hline
\end{tabular}}
\end{table}
\section{Ablation Study}
\label{Sec:ablation}
To justify our design choices, we ablate each component in this section. Due to space constraints, additional results are provided in the supplementary material.
\subsection{Effect of Collision Modeling}
\label{sec:ablationColFac}

We ablate collision modeling by setting the collision factor to $w^{\text{hash}}\equiv 1$ while keeping the same entropy-based information estimator, hash-table size, number of levels, and training protocol. \Cref{Tab:ablationcollision} shows that removing $w^{\text{hash}}$ consistently hurts reconstruction quality (PSNR/SSIM) and is less parameter-efficient, indicating that explicitly modeling collisions is a key component for allocating capacity effectively in hash-based encoders.
\subsection{Alternative Information Proxies for Resolution Scheduling}
\label{sec:ablation_info_proxy}
\begin{table}[]
\centering
\caption{Effect of alternative information proxies for resolution scheduling on Kodak. We replace the entropy-based band information in \Cref{eq:informband} with an $\ell_2$ energy proxy (Ours + $\ell_2$); variance proxy (Ours + var) and compare against the entropy proxy (Ours + entropy); entropy consistently yields higher PSNR/SSIM.}
\label{Tab:ablation:informationprox}
\resizebox{\textwidth}{!}{\begin{tabular}{ccccccccccc}
\toprule
\rowcolor{mygray} Method     & \multicolumn{2}{c}{I-NGP \cite{muller2022instant}}  & \multicolumn{2}{c}{MetricGrid \cite{wang2025metricgrids}} & \multicolumn{2}{c}{Ours+$\ell_2$} & \multicolumn{2}{c}{Ours+var} & \multicolumn{2}{c}{Ours+entropy} \\
\cmidrule(lr){2-3} \cmidrule(lr){4-5} \cmidrule(lr){6-7} \cmidrule(lr){8-9} \cmidrule(lr){10-11}
Params (K) & 206          & 300         & 207            & 335            & 214           & 334          & 211              & 331   & 205              & 332           \\ 
\textbf{PSNR} $\uparrow$      & 37.06        & 43.88       & 39.73          & 44.51          & 38.52         & 43.60        & 38.74            & 44.94    & 41.03            & 46.63         \\
\textbf{SSIM} $\uparrow$      & ~0.9386~      & ~0.9964~      & ~0.9568 ~        &~ 0.9978 ~        & ~0.9504 ~       & ~ 0.9913 ~      & ~ 0.9560 ~          & ~ 0.9893 ~   & ~ 0.9714 ~          & ~ 0.9982 ~         \\ \hline
\end{tabular}}
\end{table}

As described in \Cref{sec:methodology}, our method uses empirical entropy to quantify the effective information in each frequency band. In this ablation, we replace the entropy term in \Cref{eq:informband} with an $\ell_2$ energy proxy or variance, and rerun the Kodak experiments under comparable parameter budgets. The results in \Cref{Tab:ablation:informationprox} show a clear drop in both PSNR and SSIM when entropy is replaced by energy or variance. This behavior is expected: $\ell_2$ energy characterizes average power but is largely insensitive to the underlying distribution of the band-residual signal, and variance is likewise a poor proxy for information. In contrast, entropy directly quantifies coding uncertainty, \ie, the expected number of bits required to describe samples under an optimal code, thereby aligning with our goal of equalizing effective representational demand across levels.
\section{Conclusion}
In this paper, we presented collision-aware resolution adaptation (CARA), a data-adaptive framework for optimizing multiresolution hash encodings in implicit neural representations. By equalizing effective information across levels and explicitly accounting for collision-induced capacity reduction, CARA improved parameter utilization without increasing model size. We also introduced a pixel-shuffle strategy to further mitigate hash collisions in a parameter-efficient manner. Experiments showed consistent improvements in the fidelity--parameter trade-off across Kodak, gigapixel images, and whole-slide images, achieving comparable or better reconstruction quality with substantially fewer parameters. Limitations and future work are deferred to the supplementary material due to space constraints.


%
%
\bibliographystyle{splncs04}
\bibliography{main}




\title{CARA: Collision-Aware Resolution Adaptation for Multiresolution Hash Encoding Based Image Fitting\\{\large Supplementary Material}}

\titlerunning{CARA}

\author{Linfeng Ye\inst{1, 3}\orcidlink{0009-0009-2355-1773} \and
Zhixiang Chi\inst{1}\orcidlink{0000-0003-4560-4986} \and
Shayan Mohajer Hamidi\inst{2}\orcidlink{0000-0001-8321-7130} \and
En-hui Yang \inst{3}\orcidlink{0000-0003-1504-7584} \and 
Konstantinos N.~Plataniotis\inst{1}\orcidlink{0000-0003-3647-5473}}

\authorrunning{L.~Ye et al.}

\institute{
\mbox{$^{1}$University of Toronto \quad $^{2}$Stanford University \quad $^{3}$University of Waterloo}\\
\mbox{\email{linfeng.ye@mail.utoronto.ca; \{zhxchi;kostas\}@ece.utoronto.ca;}}\\
\mbox{\email{smohajer@stanford.edu; ehyang@uwaterloo.ca}}
}

\maketitle

\setcounter{section}{0}
\setcounter{figure}{0}
\setcounter{table}{0}
\setcounter{equation}{0}
\renewcommand{\thesection}{S\arabic{section}}
\renewcommand{\thefigure}{S\arabic{figure}}
\renewcommand{\thetable}{S\arabic{table}}
\renewcommand{\theequation}{S\arabic{equation}}


\section{Proof of Proposition 1.}
Let $U_\ell$ denote the number of distinct occupied table entries after the $n_\ell$
keys are hashed into a table of size $T$. For each table entry $t\in [T]$, define
the indicator
\begin{align}
Z_t \triangleq \mathbf{1}\{\text{entry } t \text{ is occupied}\},
\end{align}
so that
\begin{align}
U_\ell = \sum_{t=1}^{T} Z_t .
\end{align}

Under uniform hashing, each key is mapped independently and uniformly to one
of the $T$ table entries. Hence, for any fixed $t\in [T]$,
\begin{align}
\Pr(Z_t=0)=\left(1-\frac{1}{T}\right)^{n_\ell},
\qquad
\Pr(Z_t=1)=1-\left(1-\frac{1}{T}\right)^{n_\ell}.
\end{align}
Therefore,
\begin{align}
\mathbb{E}[U_\ell]
=
\sum_{t=1}^{T}\mathbb{E}[Z_t]
=
T\left(1-\left(1-\frac{1}{T}\right)^{n_\ell}\right).
\end{align}

Dividing by $n_\ell$ and using $\alpha_\ell = n_\ell/T$ yields the exact finite-table
occupancy factor
\begin{align}
\frac{\mathbb{E}[U_\ell]}{n_\ell}
=
\frac{1-\left(1-\frac{1}{T}\right)^{n_\ell}}{\alpha_\ell}.
\end{align}
Equivalently,
\begin{align}
\frac{\mathbb{E}[U_\ell]}{n_\ell}
=
\frac{1-\left(1-\frac{1}{T}\right)^{\alpha_\ell T}}{\alpha_\ell}.
\end{align}

Now consider the large-table regime with fixed load factor $\alpha_\ell$.
Using the standard limit
\begin{align}
\left(1-\frac{1}{T}\right)^{\alpha_\ell T}\to e^{-\alpha_\ell}
\qquad \text{as } T\to\infty,
\end{align}
we obtain
\begin{align}
\frac{\mathbb{E}[U_\ell]}{n_\ell}
=
\frac{1-e^{-\alpha_\ell}}{\alpha_\ell}
\triangleq
w_\ell^{\mathrm{hash}}.
\end{align}
This proves the expression in \Cref{eq:whash}.

For the second claim, suppose that the pre-collision information associated with
level $\ell$ is $B$ bits distributed across the $n_\ell$ logical keys, so that each key
carries on average $B/n_\ell$ bits. After hashing, collisions merge multiple logical
keys into the same table entry, and only $U_\ell$ distinct degrees of freedom remain
retrievable. Under this first-order occupancy model,
\begin{align}
B_{\mathrm{eff}}
=
\frac{B}{n_\ell} \, U_\ell .
\end{align}
Taking expectations and applying the result above gives
\begin{align}
\mathbb{E}[B_{\mathrm{eff}}]
=
\frac{B}{n_\ell}\,\mathbb{E}[U_\ell]
=
B\cdot \frac{\mathbb{E}[U_\ell]}{n_\ell}
=
B\cdot w_\ell^{\mathrm{hash}}.
\end{align}
In the next section, we examine the applicability of the uniform hashing assumption.

\section{On the Uniform Hashing Assumption}

Proposition~1 models collision-induced capacity reduction under the standard uniform-hashing assumption, yielding the closed-form efficiency factor
\(
w^{\mathrm{hash}}_\ell = (1-e^{-\alpha_\ell})/\alpha_\ell
\),
where \(\alpha_\ell = n_\ell/T\) denotes the load factor at level \(\ell\).
While assumption is not intended as a literal claim that the realized collision pattern of every level is exactly uniform.
It serves as a tractable first-order model of how usable capacity decreases as the number of hashed vertices grows relative to the table size.

To assess whether a more exact collision statistic is beneficial in practice, we additionally replaced the analytic factor in Proposition~1 with an empirical occupancy ratio computed from the realized hash assignments.
Specifically, for a candidate resolution \(N^{(k)}\), let \(n_k=(N^{(k)})^2\) be the number of grid vertices and let \(U_k\) denote the number of distinct occupied hash entries after mapping these vertices into a table of size \(T\).
We then define the empirical collision efficiency as
\[
\hat{w}^{\mathrm{emp}}_k \triangleq \frac{U_k}{n_k},
\]
and substitute \(\hat{w}^{\mathrm{emp}}_k\) for \(w^{\mathrm{hash}}_k\) in \Cref{eq:band_info}, while keeping all other components unchanged.

\begin{table}[]
\centering
\caption{Comparison of the uniform-hash model in CARA and an empirical occupancy-based alternative on Kodak and Tokyo. The empirical variant does not provide a consistent improvement in PSNR or SSIM across parameter budgets. This indicates that the analytic uniform-hash assumption is sufficiently accurate for collision-aware resolution scheduling, while being simpler and more tractable.}
\label{tab:uniform_hash_kodak_hash_tokyo}
\begin{tabular}{ccccccccc}
\toprule
\rowcolor{mygray} Data       & \multicolumn{4}{c}{Kodak}                                   & \multicolumn{4}{c}{Tokyo}                                   \\
\cmidrule(lr){2-5} \cmidrule(lr){6-9}
Hash Model & \multicolumn{2}{c}{Uniform} & \multicolumn{2}{c}{Empirical} & \multicolumn{2}{c}{Uniform} & \multicolumn{2}{c}{Empirical} \\
Params     & 205K        & 341K        & 206K         & 344K          & 66.34M      & 123.4M      & 66.29       & 126.64 M      \\
PSNR       & 41.03        & 46.24        & 41.07         & 46.03         & 35.01        & 37.22        & 35.04         & 37.17         \\
SSIM       & ~0.9714~       & ~0.9906~       & ~0.9713~        & ~0.9902~        & ~0.9145~       & ~0.9424~       & ~0.9163~        & ~0.9427~        \\ \hline
\end{tabular}
\end{table}

\Cref{tab:uniform_hash_kodak_hash_tokyo} report the results on Kodak and Tokyo.
In both cases, replacing the analytic uniform-hash model with the empirical occupancy estimate does not yield a consistent improvement in reconstruction quality, suggests that the dominant effect of collisions is already captured by the load factor \(\alpha_k\) under the uniform hashing assumption. In practice, moderate deviations from ideal uniform hashing only slightly perturb the per-band weights and therefore often lead to the same, or very similar, partition of the candidate resolutions. As a result, a more detailed empirical estimate of occupancy does not necessarily translate into better final PSNR or SSIM, while increasing the preprocessing time.

These results support the use of the uniform-hashing model in CARA. In particular, the uniform-hashing assumption serves as an effective surrogate for schedule construction: it preserves the essential monotonic dependence on load factor, integrates naturally into the collision-aware objective, and performs on par with empirical alternatives on both low-resolution and gigapixel images while remaining substantially simpler.

\section{Dynamic Programming Solver for CARA Resolution Scheduling}
\begin{algorithm}[t]
\caption{Dynamic programming solver}
\label{alg:cara_dp_numpy}
\begin{algorithmic}[1]
\Require Candidate resolutions $\{N^{(k)}\}_{k=1}^{K}$, number of levels $L$, hash-table size $T$, discretized information masses $\{\psi_k\}_{k=1}^{K}$
\Ensure Optimal cutoffs $0=k_0<k_1<\cdots<k_L=K$, selected resolutions $\{N_\ell\}_{\ell=1}^{L}$, optimal objective value $D[L,K]$

\State $P \gets [\,0,\ \mathrm{cumsum}(\psi_1,\dots,\psi_K)\,]$
\State $\alpha[1{:}K{+}1] \gets \big((N^{(1)})^2,\dots,(N^{(K)})^2\big)/T$
\State $w[1{:}K{+}1] \gets \bigl(1-\exp(-\alpha[1{:}K{+}1])\bigr)/ \alpha[1{:}K{+}1]$
\State $D \gets +\infty,\qquad \Pi \gets -1$
\State $D[1,1{:}K{+}1] \gets w[1{:}K{+}1] * P[1{:}K{+}1],\qquad \Pi[1,1{:}K{+}1] \gets 0$

\For{$\ell=2,\dots,L$}
    \For{$k=\ell,\dots,K$}
        \State $\mathbf{u} \gets \max\!\Big(D[\ell\!-\!1,\ell\!-\!1{:}k],\; w[k]\big(P[k]-P[\ell\!-\!1{:}k]\big)\Big)$
        \State $r \gets \arg\min \mathbf{u};\quad D[\ell,k] \gets \mathbf{u}[r];\quad \Pi[\ell,k] \gets (\ell-1)+r$
    \EndFor
\EndFor

\State $k_L \gets K$
\For{$\ell=L,\dots,2$}
    \State $k_{\ell-1} \gets \Pi[\ell,k_\ell]$
\EndFor
\For{$\ell=1,\dots,L$}
    \State $N_\ell \gets N^{(k_\ell)}$
\EndFor

\State \Return $\{k_\ell\}_{\ell=0}^{L}$, $\{N_\ell\}_{\ell=1}^{L}$, $D[L,K]$
\end{algorithmic}
\end{algorithm}

In the main paper, a resolution schedule is specified by the ordered indices
\begin{align}
0 = k_0 < k_1 < \cdots < k_L = K,
\end{align}
which partition the candidate spectrum into $L$ adjacent bands. The effective information assigned to band $\ell$ is
\begin{align}
\mathcal{I}_\ell
\triangleq
w^{\mathrm{hash}}_{k_\ell}
\sum_{f \in \mathcal{F}_\ell} \mathfrak{I}(f),
\end{align}
and the final schedule is obtained by solving the min--max problem in \Cref{eq:minmax}. Here we provide the exact dynamic programming solver used to compute this schedule.

Given the monotone candidate set of grid resolutions $\{N^{(k)}\}_{k=1}^{K}$ with
$N^{(1)} < \cdots < N^{(K)}$, we first discretize the information density
$\mathfrak{I}(f)$ over the corresponding candidate intervals. Let $\psi_k$ denote the
discretized information mass associated with the $k$-th interval, such that for any band
$(k_{\ell-1}, k_\ell]$,
\begin{align}
\sum_{j=k_{\ell-1}+1}^{k_\ell} \psi_j
=
\sum_{f \in \mathcal{F}_\ell} \mathfrak{I}(f).
\end{align}
For each candidate resolution $N^{(k)}$, the collision efficiency is
\begin{align}
w_k^{\mathrm{hash}}
\triangleq
\frac{1-e^{-\alpha_k}}{\alpha_k},
\qquad
\alpha_k \triangleq \frac{(N^{(k)})^2}{T},
\end{align}
exactly as defined in \Cref{eq:band_info}.

We define the dynamic programming state $D[\ell,k]$ as the minimum achievable value of
\(
\max_{m \in [\ell]} \mathcal{I}_m
\)
when the first $k$ candidate intervals are partitioned into $\ell$ adjacent bands.
The corresponding predecessor table is denoted by $\Pi[\ell,k]$.
The recursion is
\begin{align}
D[\ell,k]
=
\min_{j \in \{\ell-1,\dots,k-1\}}
\max\!\left(
D[\ell-1,j],\;
w_k^{\mathrm{hash}}
\sum_{t=j+1}^{k}\psi_t
\right),
\qquad
\ell=2,\dots,L,
\end{align}
with base case
\begin{align}
D[1,k]
=
w_k^{\mathrm{hash}}
\sum_{t=1}^{k}\psi_t.
\end{align}
That is, for each feasible predecessor $j$, the last band contributes the collision-aware
information
\(
w_k^{\mathrm{hash}} \sum_{t=j+1}^{k}\psi_t
\),
while the earlier bands contribute the optimal bottleneck value
\(
D[\ell-1,j]
\).
We then choose the predecessor that minimizes the resulting worst-case load.

After filling the table, the optimal cutoffs are recovered by backtracking:
we set $k_L=K$, and for $\ell=L,L-1,\dots,2$, we recursively obtain
\begin{align}
k_{\ell-1} = \Pi[\ell,k_\ell].
\end{align}
The final per-level resolutions are then
\begin{align}
N_\ell = N^{(k_\ell)}, \qquad \ell \in [L].
\end{align}

\Cref{alg:cara_dp_numpy} summarizes the resulting exact solver. Since the objective in \Cref{eq:minmax} decomposes over ordered band
partitions and exhibits optimal substructure, the algorithm solves the problem exactly
with time complexity $\mathcal{O}(LK^2)$ and memory complexity $\mathcal{O}(LK)$.

\begin{table}[]
\centering
\caption{Wall-clock training time, reconstruction quality, and parameter count of Instant-NGP, MetricGrids, and CARA on the Kodak dataset and the Tokyo gigapixel image. All methods are trained for the same number of epochs.}
\label{Tab:WallClock}
\resizebox{0.8\textwidth}{!}{\begin{tabular}{ccccccccc}
\toprule
\rowcolor{mygray} Data        & \multicolumn{4}{c}{Kodak}           & \multicolumn{4}{c}{Tokyo}           \\
\cmidrule(lr){2-5} \cmidrule(lr){6-9}
            & PSNR  & SSIM   & Time  & Params (K) ~& ~PSNR  & SSIM   & Time  & Params (M) \\
Instant-NGP \cite{muller2022instant}      & 38.70 & ~0.9386~ & 589s  & 207        & 36.26 & ~0.9281~ & 2230s & 128.0      \\
MetricGrids \cite{chi2025learning} & 39.73 & ~0.9568~ & 1722s & 207        & 35.19 & ~0.9578~ & 4180s & 128.1      \\
Ours        & 41.03 & ~0.9714~ & 592s  & 205        & 37.22 & ~0.9424~ & 2252s & 123.4      \\ \hline
\end{tabular}}
\vspace{-1cm}
\end{table}

\section{Implementation Details and Wall-Clock Cost}
Our implementation of CARA follows the same training pipeline as the hybrid
INR baselines, and largely follows the MetricGrid setting whenever applicable.
The only method-specific modifications are the collision-aware resolution
schedule and the optional pixel-shuffle preprocessing. Unless otherwise
specified, all models are optimized with Adam using
$(\beta_1,\beta_2,\epsilon)=(0.9,0.99,10^{-15})$, cosine annealing learning-rate
decay, and an $\ell_2$ reconstruction loss.

Across all experiments, we use the same multiresolution hash-grid backbone for
all hybrid methods. The feature dimension is fixed to $F=2$ at every level, and
the number of hash-grid levels and decoder architecture are kept unchanged across methods and parameter budgets. To match different parameter budgets while keeping the architecture depth fixed, we vary only the hash-table size $T$.

For Kodak, we follow the MetricGrid training protocol and train all methods for
20{,}000 optimization steps under comparable parameter budgets. As in the main
paper, pixel shuffle is not applied to Kodak owing to its relatively low
resolution. For gigapixel natural images and WSIs, we use the same optimizer
settings.

We also report the per-experiment wall-clock training time on \emph{Nvidia V100} GPUs, in \Cref{Tab:WallClock}. Since CARA only adaptively assigns the resolution schedule while leaving the underlying model architecture unchanged, it incurs only a negligible increase in training time relative to Instant-NGP, while achieving the best reconstruction fidelity. In contrast, MetricGrids relies on a more complex decoder, which substantially increases the training time.

\section{Gigapixel Image Results}
\begin{table}[]
\centering
\caption{Numerical comparison on gigapixel Pluto image fitting. We report PSNR (dB), SSIM and the corresponding number of trainable parameters (in millions). Our method maintains competitive fidelity in both the low and high-parameter regimes.}
\label{tab:PlutoResults}
\resizebox{\textwidth}{!}{
\begin{tabular}{cccccccccccc}
\toprule
\rowcolor{mygray} Method     & \multicolumn{3}{c}{~~I-NGP \cite{muller2022instant}~~} & ~~NeuRBF \cite{chen2023neurbf}~~  & \multicolumn{3}{c}{~~MetricGrid \cite{wang2025metricgrids}~~} & \multicolumn{4}{c}{Ours}       \\ 
\cmidrule(lr){2-4} \cmidrule(lr){5-5}\cmidrule(lr){6-8}\cmidrule(lr){9-12} 
                   Params (M) &  16.01  &  64.01  & 217.3 &  57.46  &  8.02  & 128.02     & 217.4   & 7.89  &  48.07 & 108.23 ~ &  202.9 \\
\textbf{PSNR} $\uparrow$      &  40.37  &  54.66   & 61.54   & 53.35   & 41.51 & 61.2   & 63.51    & 42.52 &  54.94   &  64.32 ~  &  65.56 \\ 
\textbf{SSIM} $\uparrow$      & ~0.9584~ & ~0.9983~   & ~0.9996~  &  0.9924  & ~0.9585~  & ~0.9996~  & ~0.9998~    & ~0.9778~ & ~ 0.9873 ~ &  ~0.9999~  & ~0.9999~  \\ \hline
\end{tabular}}
\end{table}

\begin{figure}[]
\centering
\includegraphics[width=\linewidth]{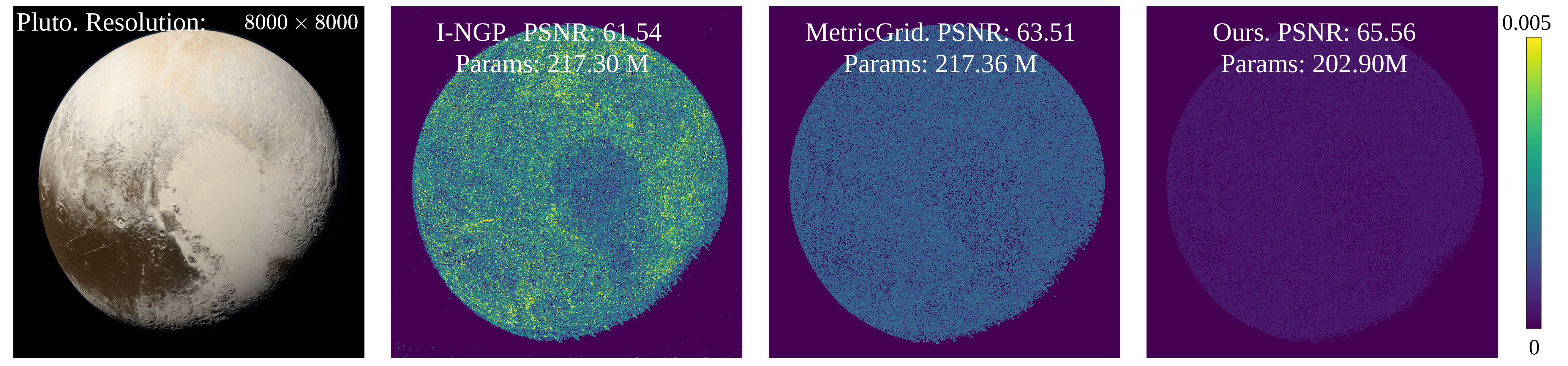}
\caption{Qualitative comparison on the Pluto gigapixel image. The left panel presents the ground-truth image, while the remaining panels show per-pixel $\ell_2$ error maps for Instant-NGP, MetricGrid, and the proposed method, respectively. Brighter colors correspond to larger reconstruction error. Although MetricGrid improves over Instant-NGP on this example, our method yields the lowest residual error overall and achieves the highest PSNR (65.56 dB) while using fewer parameters, indicating a superior fidelity--parameter trade-off.}
\label{Fig:pluto_res}
\end{figure}
We present the full quantitative and qualitative results on the remaining two gigapixel scenes, \emph{Tokyo} and \emph{Pluto}, complementing the \emph{Girl} results reported in the main paper. These two images stress the resolution schedule in different ways. \emph{Tokyo} contains dense, spatially varying high-frequency structures, whereas \emph{Pluto} is smoother at the global scale but still requires accurate modeling of subtle multi-scale boundaries and texture variations. Together, they provide a useful test bed for evaluating whether a resolution schedule generalizes across different gigapixel images.

\Cref{tab:PlutoResults} shows that CARA also yields consistent gains on \emph{Pluto}. In the low-parameter regime, CARA attains 42.52\,dB with 7.89M parameters, outperforming both I-NGP (40.37\,dB, 16.01M) and MetricGrid (41.51\,dB, 8.02M). In the high-fidelity regime, CARA reaches 65.56\,dB with 202.9M parameters, surpassing MetricGrid's 63.51\,dB at 217.4M parameters and I-NGP's 61.54\,dB at 217.3M parameters. The residual maps in \Cref{Fig:pluto_res} are consistent with the quantitative results: CARA produces the smallest high-error regions and the lowest overall reconstruction error. Unlike \emph{Tokyo}, where the weakness of non-adaptive schedules is especially pronounced, \emph{Pluto} shows that even when a strong baseline performs well, collision-aware adaptation still leads to a better fidelity--parameter trade-off.

\begin{table}[]
\centering
\vspace{-0.5cm}
\caption{Numerical comparison on gigapixel Tokyo image fitting. We report PSNR (dB), SSIM and the corresponding number of trainable parameters (in millions). Our method maintains competitive fidelity in both the low and high-parameter regimes.}
\label{tab:TokyoResults}
\resizebox{\textwidth}{!}{
\begin{tabular}{cccccccccccc}
\toprule
\rowcolor{mygray} Method     & \multicolumn{3}{c}{~~I-NGP \cite{muller2022instant}~~} & ~~NeuRBF \cite{chen2023neurbf}~~  & \multicolumn{3}{c}{~~MetricGrid \cite{wang2025metricgrids}~~} & \multicolumn{4}{c}{Ours}       \\ 
\cmidrule(lr){2-4} \cmidrule(lr){5-5}\cmidrule(lr){6-8}\cmidrule(lr){9-12} 
                   Params (M) & 8.01      &   128.0~    & ~442.4   & 131.7  & 16.01     & 128.1~     & 442.6     & ~6.55~  & 66.34~ & 123.4~ & 468.0 \\
\textbf{PSNR} $\uparrow$      & 18.56     &   36.26    & ~38.45   & 34.32  & 24.72    & ~35.19 ~   & ~37.75~    & ~24.64~ &  35.01~  &  37.22~  & 41.85 \\ 
\textbf{SSIM} $\uparrow$      & ~ 0.3212 ~ & ~ 0.9281 ~ & ~0.9362 & 0.8936  & ~0.8173~  & ~0.9578 ~   & ~0.9423~    & ~0.8424~ & ~0.9145 ~ &  0.9424~  & 0.9916 \\ \hline
\end{tabular}}
\end{table}
\begin{figure}[]
\centering
\includegraphics[width=\linewidth]{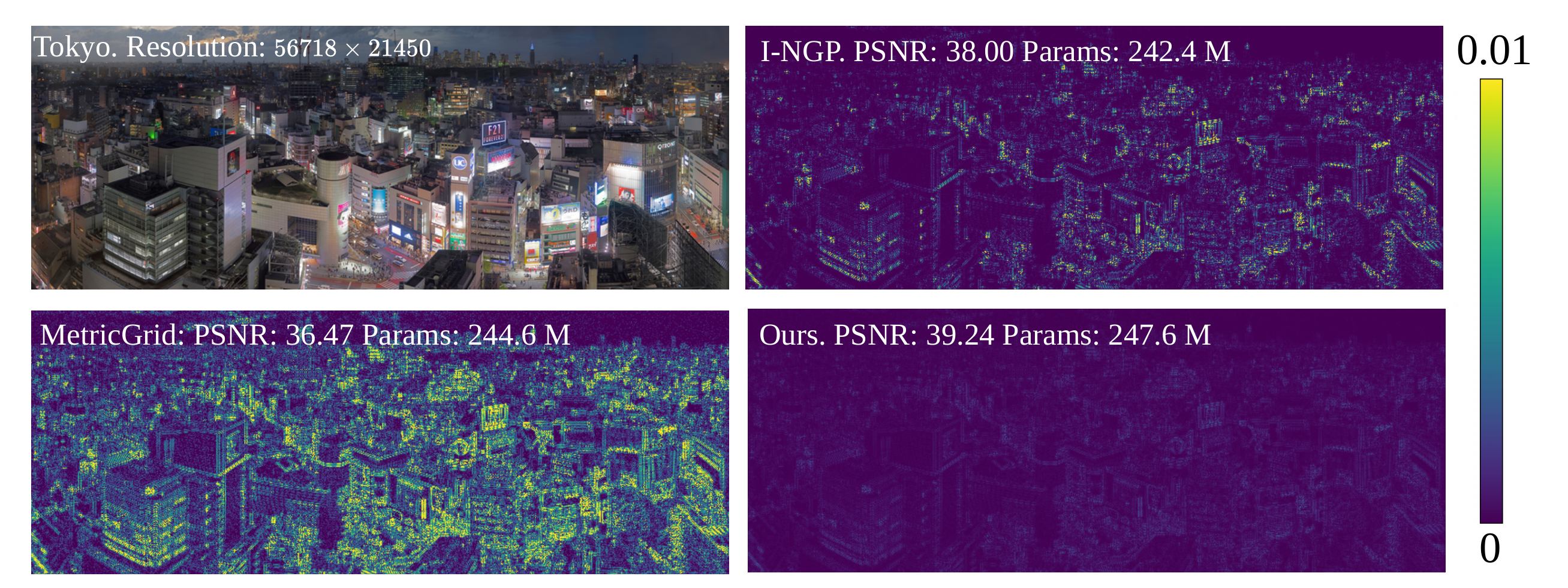}
\caption{Qualitative comparison on the Tokyo gigapixel image. The left-up panel shows the ground-truth image, and the remaining panels visualize per-pixel $\ell_2$ error maps for Instant-NGP, MetricGrid, and the proposed method, respectively. Brighter colors indicate larger reconstruction error. Notably, on this image, MetricGrid underperforms Instant-NGP despite using a comparable parameter budget, yielding both a lower PSNR (36.47 dB vs.\ 38.00 dB) and visibly larger high-error regions. In contrast, our method produces the lowest residual error and the highest reconstruction fidelity (39.24 dB), demonstrating the benefit of collision-aware, data-adaptive resolution allocation on challenging gigapixel scenes.}
\label{Fig:tokyo_res}
\end{figure}

\Cref{tab:TokyoResults} shows that \emph{Tokyo} is a particularly challenging case for data-agnostic or hand-crafted schedules. In the low-parameter regime, CARA achieves 24.64\,dB using only 6.55M parameters, which is comparable to MetricGrid's 24.72\,dB obtained with 16.01M parameters, while substantially outperforming I-NGP (21.51\,dB). As the parameter budget increases, the benefit of data-adaptive allocation becomes even clearer: CARA is the only method in \Cref{tab:TokyoResults} to exceed 40\,dB, reaching 41.85\,dB. The qualitative comparison in \Cref{Fig:tokyo_res} further highlights this effect. At a comparable parameter budget, MetricGrid underperforms even I-NGP on this image, yielding larger high-error regions, whereas CARA produces the lowest residual error map and the highest reconstruction fidelity. This behavior indicates that fixed or heuristic schedules can be brittle on scenes with highly non-uniform frequency content, and highlights the necessity of adapting the per-level resolutions to the input image.

Taken together with the \emph{Girl} results in the main paper, these experiments show that the optimal resolution allocation is strongly image-dependent. A data-agnostic schedule may be competitive on some scenes, but it is not reliably robust across gigapixel images. This is precisely where CARA becomes necessary. By selecting the per-level resolutions through collision-aware effective information balancing, CARA avoids overloading some levels while under-utilizing others, leading to more efficient use of parameters and consistently higher reconstruction fidelity across diverse gigapixel content.

\section{WSI Data Curation and Details}
\begin{figure}[]
\centering
\includegraphics[width=0.85\linewidth]{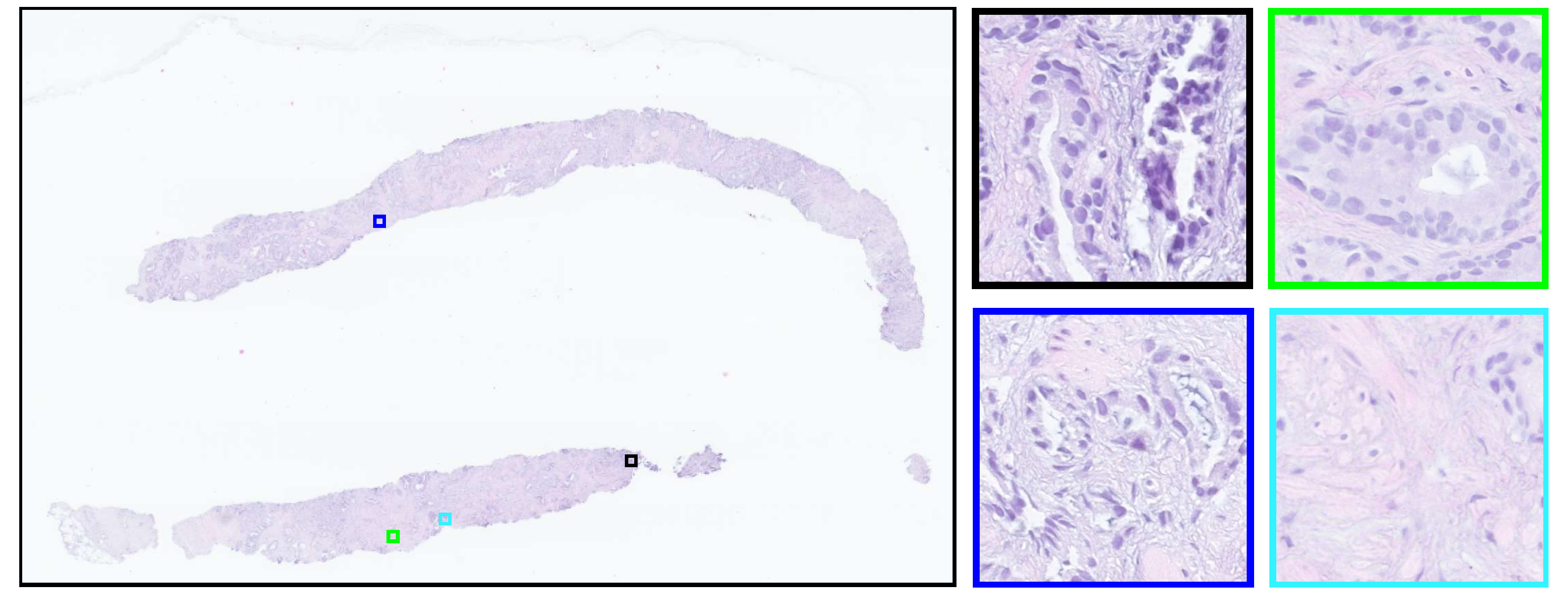}
\caption{Slide 1 and selected patch regions. Left: downsampled WSIs, where colored boxes indicate the spatial locations of the sampled patches. Right: the corresponding high-resolution selected patches extracted from the marked regions.}
\label{Fig:WSISamples1}
\end{figure}
We collected four de-identified $H\&E$-stained whole-slide images (WSIs) at 40$\times$ magnification through {Huron Technologies} using a {TissueScope whole-slide} scanner. The acquisition workflow consists of an initial setup stage, including preview scanning, verification, and batch pre-processing, followed by full-resolution whole-slide digitization with the {TissueScope} scanner. For this study, the scanned slides were directly exported in BigTIFF format, and the uncompressed exports were retained for all experiments.
\begin{figure}[]
\centering
\includegraphics[width=0.85\linewidth]{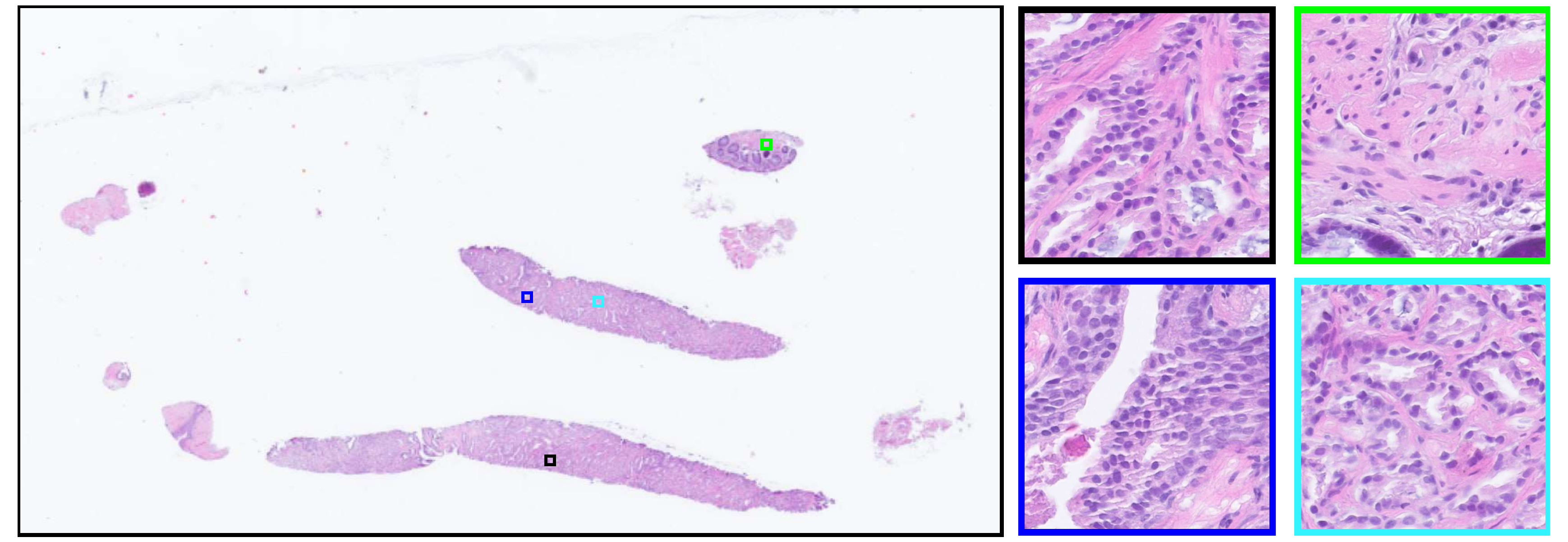}
\caption{Slide 2 and selected patch regions. Left: downsampled WSIs, where colored boxes indicate the spatial locations of the sampled patches. Right: the corresponding high-resolution selected patches extracted from the marked regions.}
\label{Fig:WSISamples2}
\end{figure}
To maximize image fidelity and avoid confounding effects from downstream enhancement or compression, we retained only gamma correction, and performed no additional preprocessing after export. In particular, we did not apply stain normalization, denoising, deblurring, sharpening, contrast manipulation, or lossy compression. Hence, in this supplementary material, \emph{raw WSI} denotes the direct scanner-exported TIFF data after minimal scanner-side conversion, but before any further computational preprocessing. All patient-identifiable information was removed before use.

\begin{figure}[]
\centering
\includegraphics[width=0.85\linewidth]{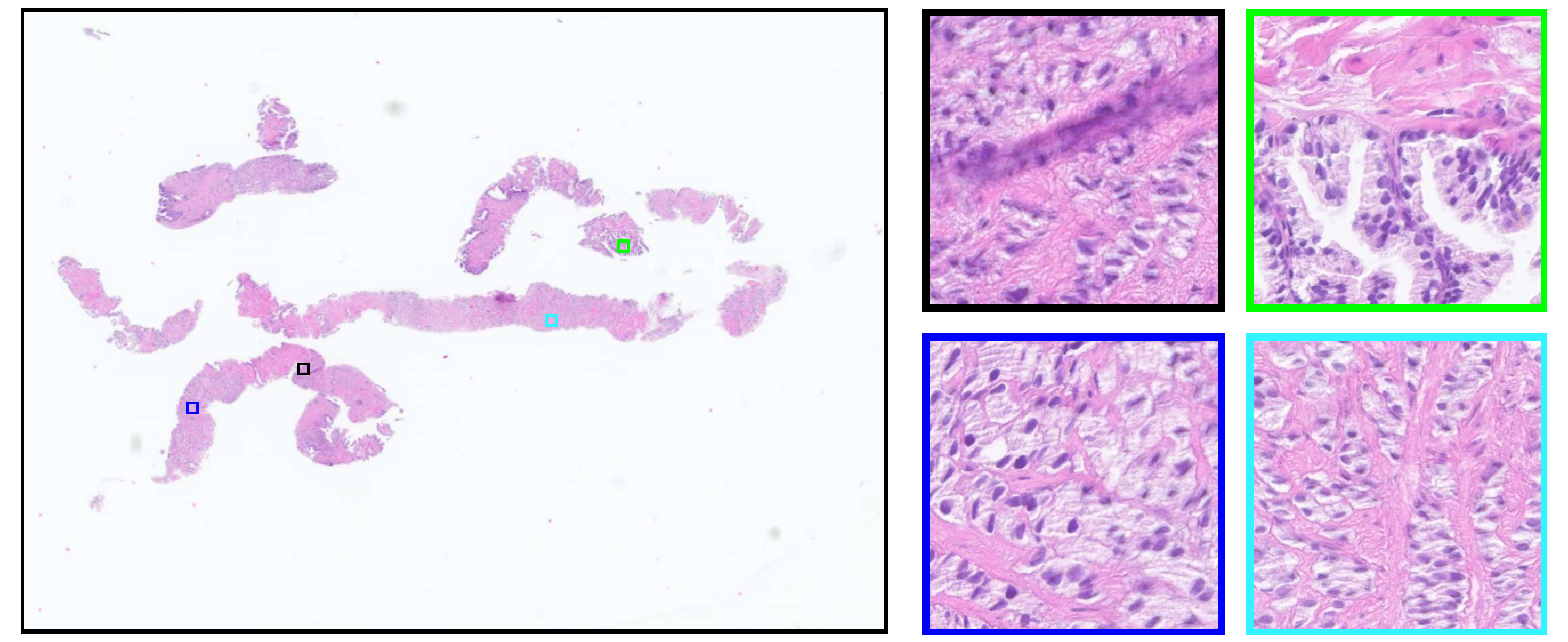}
\caption{Slide 3 and selected patch regions. Left: downsampled WSIs, where colored boxes indicate the spatial locations of the sampled patches. Right: the corresponding high-resolution selected patches extracted from the marked regions.}
\label{Fig:WSISamples3}
\end{figure}

\begin{figure}[]
\centering
\includegraphics[width=0.85\linewidth]{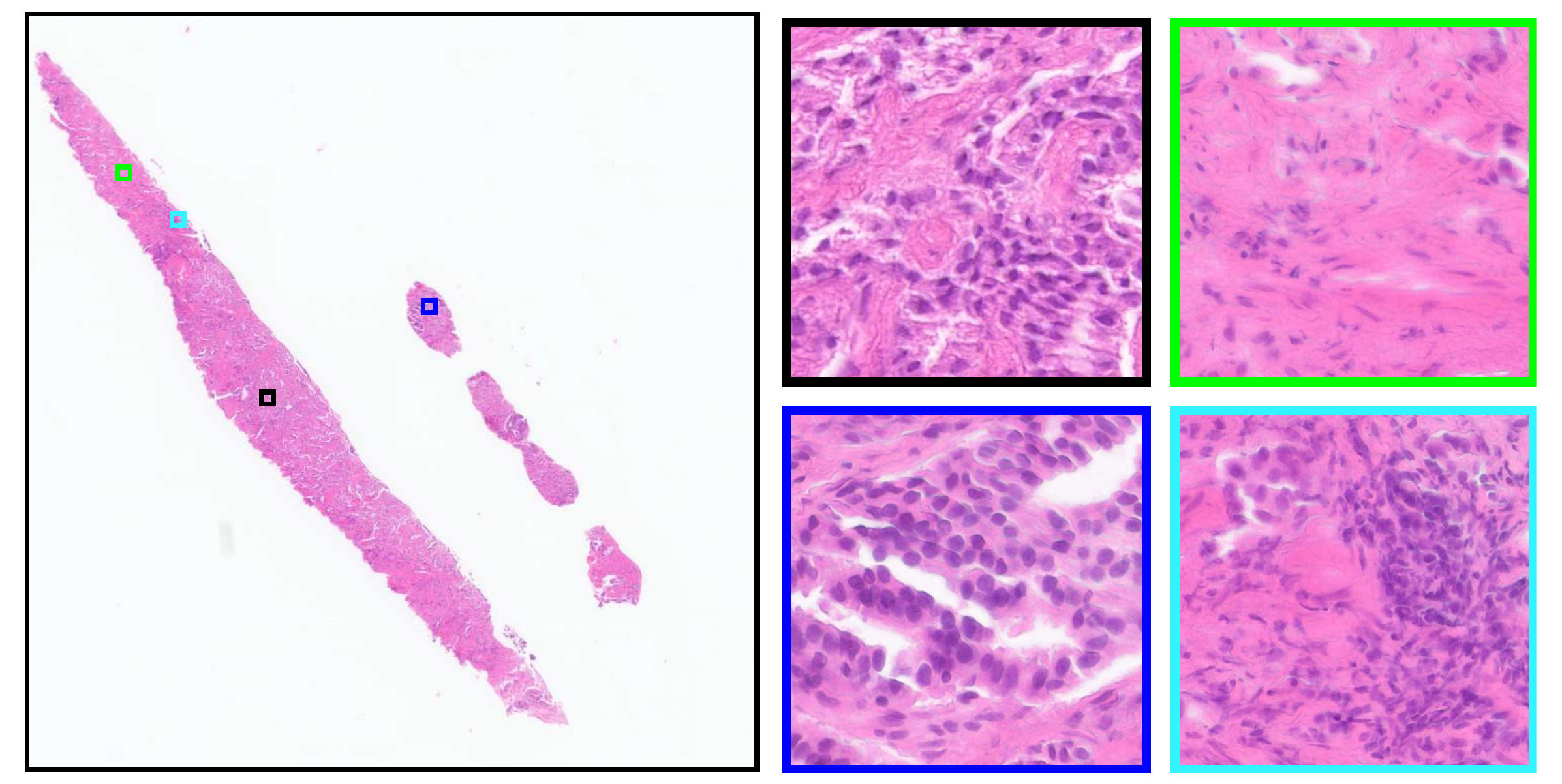}
\caption{Slide 4 and selected patch regions. Left: downsampled WSIs, where colored boxes indicate the spatial locations of the sampled patches. Right: the corresponding high-resolution selected patches extracted from the marked regions.}
\label{Fig:WSISamples4}
\end{figure}

Downsampled WSIs and their corresponding selected patch samples are shown in \Cref{Fig:WSISamples1,Fig:WSISamples2,Fig:WSISamples3,Fig:WSISamples4}. 

\section{Ablation Study}

\subsection{Ablation of the Low-Pass Filter Choice}
\label{sec:supp_lowpass_ablation}

As defined in \Cref{eq:informband}, CARA estimates the band information density from the low-pass sequence $\mathbf{Y}_{\le f}$ through
\[
\mathfrak{I}(f)=H\!\left(\mathbf{Y}_{\le f+\Delta f}-\mathbf{Y}_{\le f}\right)\cdot N_f^2.
\]
Therefore, the choice of low-pass filter can affect the estimated information distribution across frequency bands, and hence the final resolution schedule selected by CARA. In the main paper, we use anti-aliased resizing as the default low-pass operator. Here, we justify this choice by comparing Gaussian filtering, anti-aliased resizing, and Butterworth filtering.

\begin{table}[]
\centering
\caption{Ablation of the low-pass filter used to estimate the band information density for CARA. We compare Gaussian filtering, Butterworth filtering, and anti-aliased resizing on Kodak and Pluto. \textbf{Bold} and \underline{underlined} values denote the best and second-best results.} \label{tab:lowpass_kodak_pluto}
\begin{tabular}{ccccccc}
\toprule
\rowcolor{mygray} Data   & \multicolumn{3}{c}{Kodak}      & \multicolumn{3}{c}{Pluto}   \\ \cmidrule(lr){2-4} \cmidrule(lr){5-7}
Filter & Gaussian & Butterworth & Resize ~& Gaussian & Butterworth & Resize   \\
Params & 206.3 K  & 206 K       & 205 K ~ & 112.46 M & OOM         & 108.23 M \\
\textbf{PSNR} $\uparrow$   & 40.07    & \textbf{41.22}       & \underline{41.03} ~ & \underline{62.52}    & OOM         & \textbf{64.32}    \\
\textbf{SSIM} $\uparrow$   & 0.9576   & \textbf{0.9732}      & \underline{0.9714} ~ & \underline{0.9999}   & OOM         & \textbf{0.9999}   \\ \hline
\end{tabular}
\end{table}

Results on Kodak and the gigapixel Pluto image are reported in \Cref{tab:lowpass_kodak_pluto}. A consistent trend is that Gaussian filtering performs the worst among the tested choices. This is observed on both Kodak and Pluto, suggesting that the corresponding low-pass sequence is less suitable for estimating the band information profile used by CARA, which in turn leads to inferior reconstruction performance. On Kodak, Butterworth filtering with $K=2$ achieves the best result, while anti-aliased resizing remains very close. This indicates that CARA is not overly sensitive to the exact low-pass operator, as long as the filter provides a reliable coarse-to-fine decomposition for estimating $\mathfrak{I}(f)$. In particular, the small gap between Butterworth filtering and anti-aliased resizing on Kodak suggests that resize-based low-pass filtering is already sufficient to recover a high-quality resolution schedule.

However, Butterworth filtering is much less practical for ultra-high-resolution inputs. In our implementation, it operates in the frequency domain and requires Fourier transforms over the full image, which leads to prohibitive memory usage and out-of-memory (OOM) errors on gigapixel images. As a result, we could not evaluate Butterworth filtering on Pluto and other ultra-high-resolution examples. By contrast, anti-aliased resizing scales well to gigapixel inputs, performs on par with Butterworth on Kodak, and substantially outperforms Gaussian filtering on Pluto.

Taken together, these results support the use of anti-aliased resizing as the default low-pass filter in all experiments. Although Butterworth filtering can provide a slight advantage on a moderate-resolution image such as Kodak, it does not scale to the gigapixel regime considered in this work. Anti-aliased resizing offers the best overall trade-off between reconstruction quality and scalability, while remaining clearly more reliable than Gaussian filtering. Therefore, all experiments in the main paper use anti-aliased resizing to construct $\mathbf{Y}_{\le f}$ and compute $\mathfrak{I}(f)$.

\begin{table}[]
\centering
\caption{Ablation of the pixel-shuffle folding factor $s$ on the Tokyo gigapixel image across four comparable parameter budgets. We evaluate $s \in \{1,2,4\}$, where $s=1$ denotes no pixel shuffle. Moderate shuffling ($s=2$) is most effective in the compact regimes, improving PSNR from $18.69$ to $24.64$ dB at $\sim$6.6M parameters and from $26.38$ to $31.76$ dB at $\sim$24M parameters. In contrast, at larger budgets ($\sim$91--93M and $\sim$239--248M parameters), the best performance is achieved without shuffling ($s=1$), while aggressive shuffling ($s=4$) is consistently inferior. These results show that pixel shuffle is a budget-dependent collision-mitigation strategy: it is beneficial when hash collisions are severe, but its advantage diminishes once the model has sufficient capacity.}
\label{tab:pixel_shuffle_budget}
\resizebox{\textwidth}{!}{\begin{tabular}{ccccccccccccc}
\toprule
\rowcolor{mygray}\multicolumn{13}{c}{Tokyo Gigapixel Image} \\ 
Shuffle Factor & \textbf{PSNR} $\uparrow$ & \textbf{SSIM} $\uparrow$  & Params (M) & \textbf{PSNR} $\uparrow$ & \textbf{SSIM} $\uparrow$  & Params (M) & \textbf{PSNR} $\uparrow$ & \textbf{SSIM} $\uparrow$  & Params (M) & \textbf{PSNR} $\uparrow$ & \textbf{SSIM} $\uparrow$  & Params (M) \\
\cmidrule(lr){2-4} \cmidrule(lr){5-7} \cmidrule(lr){8-10} \cmidrule(lr){11-13}
$s=1$            & 18.69 & 0.3423 & 6.568      & 26.38 & 0.8134 & 25.39      & \textbf{35.52} & \textbf{0.9274} & \textbf{92.95}      & \textbf{39.73} & \textbf{0.9612} & \textbf{247.6}      \\
$s=2$            & \textbf{24.64} & \textbf{0.8424} & \textbf{6.599}      & \textbf{31.76} & \textbf{0.9091} & \textbf{23.93}      & 34.66 & 0.9024 & 91.11      & 37.74 & 0.9434 & 238.8      \\
$s=4$            & 18.21 & 0.4462 & 6.443      & 29.63 & 0.8612 & 22.42      & 32.52 & 0.8903 & 91.02      & 36.27 & 0.9225 & 236.4      \\ \hline
\end{tabular}}
\end{table}

\subsection{Effect of Pixel Shuffle Across Parameter Budgets}

In \Cref{sec:MethodPixelShuffleing} of the main paper, pixel shuffle is introduced as a collision-mitigation mechanism for multiresolution hash encoding. For a shuffle factor $s$, the spatial resolution is reduced by a factor of $s$ along each axis, so the load factor at each hash-grid level decreases from $\alpha_\ell$ to $\alpha_\ell / s^2$. Consequently, the potential benefit of pixel shuffle should depend on the parameter regime: when the hash tables are heavily loaded, reducing collisions can noticeably improve the effective capacity of the encoder, whereas this advantage is expected to diminish once the model already has sufficient capacity.

To evaluate this effect, we ablate the folding factor $s \in \{1,2,4\}$ on the Tokyo gigapixel image across four comparable parameter budgets, where $s=1$ corresponds to the original formulation without pixel shuffle. For each setting, the shuffle transform is applied before both schedule construction and INR training, and the full-resolution prediction is recovered by the inverse transform at test time. Across all experiments, we keep the backbone architecture, number of levels, feature dimension, optimizer, and training iterations unchanged, and vary only the hash-table size to obtain different parameter regimes.

The results in \Cref{tab:pixel_shuffle_budget} show that the effect of pixel shuffle is strongly budget-dependent. In the compact regimes, moderate shuffling with $s=2$ gives the best performance, improving PSNR from $18.69$ to $24.64$ dB at approximately $6.6$M parameters and from $26.38$ to $31.76$ dB at approximately $24$M parameters, while also improving SSIM in both cases. These gains are consistent with the collision analysis in Section~4.2: when the load factor is high, reducing collisions materially improves usable capacity.

However, this trend does not persist at larger budgets. At approximately $91$--$93$M parameters, the best result is obtained with no shuffle ($s=1$), which outperforms $s=2$ by $0.86$ dB PSNR and $s=4$ by $3.00$ dB. The same behavior becomes even clearer in the highest-budget regime ($\approx 239$--$248$M parameters), where $s=1$ again yields the best reconstruction quality, exceeding $s=2$ by $1.99$ dB and $s=4$ by $3.46$ dB. This indicates that once collisions are no longer the dominant bottleneck, the advantage of shuffling saturates, while the decoder must still predict a higher-dimensional output at each queried coordinate.

In short, \Cref{tab:pixel_shuffle_budget} suggests that pixel shuffle should be viewed as an optional collision-mitigation strategy for compact ultra-high-resolution models rather than a universally optimal design choice. Moderate shuffling can substantially improve the fidelity--parameter trade-off when the encoder is collision-limited, but aggressive shuffling is consistently suboptimal in our experiments, and even moderate shuffling becomes unnecessary once sufficient model capacity is available.

\section{Limitations and Future Work}

This work focuses on the encoder side of multiresolution hash-based implicit neural representations by adapting the per-level resolution schedule to the input image, while keeping the decoder architecture fixed. This design isolates the effect of collision-aware resolution allocation and preserves compatibility with the lightweight decoders commonly used in hash-grid INRs. At the same time, it means that CARA does not yet explore joint encoder--decoder co-design, which may provide additional gains by allowing the decoder to better exploit the improved multiscale representations produced by the adaptive schedule.


A promising direction for future work is to combine CARA with more expressive yet efficient decoders, so that adaptive resolution allocation and decoding capacity can be optimized jointly within a unified INR design. Another important direction is to develop spatially aware alternatives to pixel shuffle that incorporate local image structure more explicitly while preserving computational and memory efficiency. We believe these extensions could further strengthen compact and high-fidelity implicit image representations.

\section{Ethics Statement}
The full-resolution WSIs used in this work will be publicly released upon acceptance. Due to the extremely large file size of the raw, uncompressed files, we provide subsampled WSIs and representative patches in the supplementary material. The use of these data complies with established standards for academic research. To protect privacy, all personally identifiable information and sensitive patient data have been strictly removed.

\section{LLM Usage Statement}
A large language model was used solely to polish the writing of this manuscript, such as correcting grammar, refining phrasing, and improving clarity. All core ideas, methods, experiments, results, and interpretations are entirely the authors’ own. 

\end{document}